\documentclass[10pt,twocolumn,letterpaper]{article}

\usepackage{wacv}
\usepackage{times}
\usepackage{epsfig}
\usepackage{graphicx}
\usepackage{amsmath}
\usepackage{amssymb}
\usepackage{booktabs}

\usepackage{caption}
\usepackage{ulem}
\usepackage{amsfonts}       
\usepackage{url}
\usepackage{enumitem}
\usepackage{booktabs}       
\usepackage{xcolor}
\usepackage{xspace}
\usepackage{wrapfig}
\usepackage{mathtools}
\usepackage{multirow, makecell}
\usepackage{nicefrac}       
\usepackage{pifont}
\usepackage[accsupp]{axessibility}

\newcommand{\model}{GuidingPainter\xspace}
\newcommand{\newmetric}{NRI\xspace}
\newcommand{\segnet}{segmentation network\xspace}
\newcommand{\colornet}{colorization network\xspace}
\newcommand{\hintgen}{hint generation function\xspace}
\newcommand{\compseg}{active-guidance module\xspace}
\newcommand{\compcolor}{colorization module\xspace}

\newcommand{\xmark}{\ding{55}}
\newcommand{\cmark}{\ding{51}}

\def\wacvPaperID{1448} 

\wacvalgorithmstrack   

\wacvfinalcopy 

\ifwacvfinal
\usepackage[breaklinks=true,bookmarks=false]{hyperref}
\else
\usepackage[pagebackref=true,breaklinks=true,colorlinks,bookmarks=false]{hyperref}
\fi

\begin{document}

\title{Guiding Users to Where to Give Color Hints for Efficient Interactive Sketch Colorization via Unsupervised Region Prioritization}

\newcommand*{\affaddr}[1]{#1}
\newcommand*{\affmark}[1][*]{\textsuperscript{#1}}
\newcommand*{\email}[1]{\texttt{#1}}
\author{
Youngin Cho*\affmark[1]\quad Junsoo Lee*\affmark[2]\quad Soyoung Yang\affmark[1]\quad Juntae Kim\affmark[3]\quad Yeojeong Park\affmark[1]\quad Haneol Lee\affmark[4]\\ Mohammad Azam Khan\affmark[1]\quad Daesik Kim\affmark[2]\quad Jaegul Choo\affmark[1]\\
\affaddr{\affmark[1]KAIST AI}\quad \affaddr{\affmark[2]NAVER WEBTOON AI}\quad \affaddr{\affmark[3]Korea University}\quad \affaddr{\affmark[4]UNIST}\\
\small\email{\{choyi0521,sy\_yang,indigopyj,azamkhan,jchoo\}@kaist.ac.kr}\\
\small\email{\{junsoolee93,daesik.kim\}@webtoonscorp.com}\\
\small\email{kjt7889@korea.ac.kr}\\
\small\email{haneollee@unist.ac.kr}
}

\twocolumn[{%
\renewcommand\twocolumn[1][]{#1}%
\maketitle
\centering
\vspace{-0.2in}
\includegraphics[width=\linewidth]{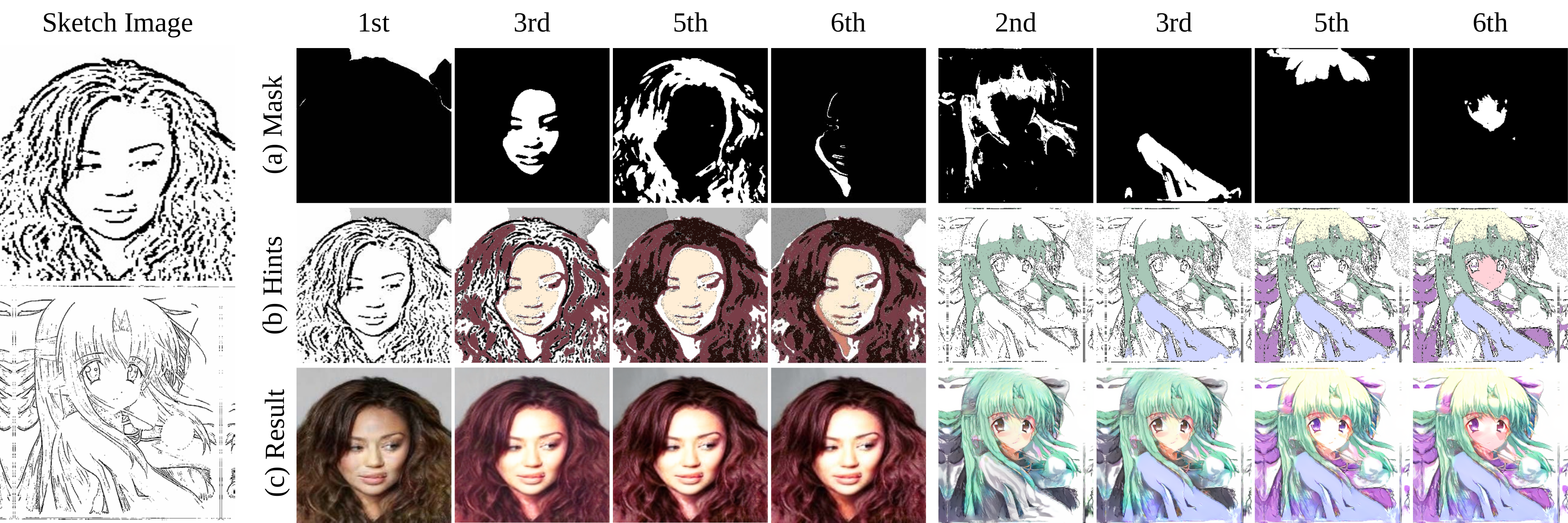}
\vspace{-0.25in}
\captionof{figure}{
\textbf{Results of our proposed model on human faces and comics datasets.} 
Each column of (a)-(c) indicates the order of interactions as the $i$-th priority.
(a) visualizes masked regions which our model guides at the $i$-th step.
Given a region as (a), users select its representative color, and the region is filled with the selected color. (c) shows intermediate colorization results for given accumulated color hints as (b).
}
\label{fig:front_door}
}]

\def\thefootnote{*}\footnotetext{Equal contribution}\def\thefootnote{\arabic{footnote}}
\thispagestyle{empty}

\begin{abstract}
Existing deep interactive colorization models have focused on ways to utilize various types of interactions, such as point-wise color hints, scribbles, or natural-language texts, as methods to reflect a user's intent at runtime. However, another approach, which actively informs the user of the most effective regions to give hints for sketch image colorization, has been under-explored. This paper proposes a novel model-guided deep interactive colorization framework that reduces the required amount of user interactions, by prioritizing the regions in a colorization model. Our method, called GuidingPainter, prioritizes these regions where the model most needs a color hint, rather than just relying on the user's manual decision on where to give a color hint. 
In our extensive experiments, we show that our approach outperforms existing interactive colorization methods in terms of the conventional metrics, such as PSNR and FID, and reduces required amount of interactions.
\end{abstract}

\vspace{-0.2in}
\section{Introduction}
The colorization task in computer vision has received considerable attention recently, since it can be widely applied in content creation. 
Most content creation starts with drawn or sketch images, and these can be accomplished within a reasonable amount of time, but fully colorizing them is a labor-intensive task. 
For this reason, the ability to automatically colorize sketch images has significant potential values.
However, automatic sketch image colorization is still challenging for the following reasons.
(i) The information provided by an input sketch image is extremely limited compared to colored images or even gray-scale ones, 
and (ii) there can be multiple possible outcomes for a given sketch image without any conditional input, 
which tends to degrade the model performance and introduce bias toward the dominant colors in the dataset.

To alleviate these issues, conditional image colorization methods take partial hints in addition to the input image, and attempt to generate a realistic output image that reflects the context of the given hints. 
Several studies have leveraged \textit{user-guided} interactions as a form of user-given conditions to the model, assuming that the users would provide a desired color value for a region as a type of point-wise color hint~\cite{zhang2017realtime} or a scribble~\cite{sangkloy2017scribbler, ci2018alacgan}.
Although these approaches have made remarkable progress, there still exist nontrivial limitations.
First, existing approaches do not address the issue of estimating semantic regions which indicate how far the user-given color hints should be spread, and thus the colorization model tends to require lots of user hints to produce a desirable output. 
Second, for every interaction at test time, the users are still expected to provide a local-position information of color hint by pointing out the region of interest (RoI), which increases the user's effort and time commitment. 
Lastly, since existing approaches typically obtain the color hints on randomized locations at training time, the discrepancies among intervention mechanisms for the training and the test phases need to be addressed. 

In this work, we propose a novel \textit{model-guided} framework for the interactive colorization of a sketch image, called \model.
A key idea behind our work is \textbf{to make a model actively seek for regions where color hints would be provided}, 
which can significantly improve the efficiency of interactive colorization process.
To this end, \model consists of two modules: \compseg and \compcolor. 
Although \compcolor works similar to previous methods, our main contribution is a hint generation mechanism in \compseg. 
The \compseg (Section~\ref{sec:method_segment-guidence}-\ref{sec:method_hint_generation}) 
(i) divides the input image into multiple semantic regions and (ii) ranks them in decreasing order of estimated model gains when the region is colorized (Fig.~\ref{fig:front_door}(a)). 

Since it is extremely expensive to obtain groundtruth for segmentation labels or even their prioritization, we explore a simple yet effective approach that identifies the meaningful regions in an order of their priority \textit{without any manually annotated labels}. 
In our active guidance mechanism (Section~\ref{sec:method_hint_generation}), \model can learn such regions by intentionally differentiating the frequency of usage for each channel obtained from the segmentation network. 
Also, we conduct a toy experiment (Section~\ref{sec:exp_segment-guide_module}) to understand the mechanism, and to verify the validity of our approach.
We propose several loss terms, e.g. smoothness loss and total variance loss, to improve colorization quality in our framework (Section~\ref{sec:method_objective_functions}), and analyze its effectiveness for both quantitatively and qualitatively (Section~\ref{sec:exp_loss_functions}). 
Note that the only action required of users in our framework is to select one representative color for each region the model provides based on the estimated priorities (Fig.~\ref{fig:front_door}(b)). 
Afterwards, the \colornet (Section~\ref{sec:method_colorization_networks}) generates a high-quality colorized output by taking the given sketch image and the color hints (Fig.~\ref{fig:front_door}(c)).

In summary, our contributions are threefold:
\begin{itemize}
  \setlength\itemsep{0.0em}
    \item We propose a novel model-guided deep image colorization framework, which 
    prioritizes regions of a sketch image in the order of the interest of the colorization model.
    \item \model can learn to discover meaningful regions for colorization and arrange them in their priority just by using the groundtruth colorized image, without additional manual supervision. 
    \item  We demonstrate that our framework can be applied to a variety of datasets by comparing it against previous interactive colorization approaches in terms of various metrics, including our proposed evaluation protocol.
\end{itemize}

\section{Related Work}
\subsection{Deep Image Colorization}
Existing deep image colorization methods, which utilize deep neural networks for colorization, can be divided into automatic and conditional approaches, depending on whether conditions are involved or not. 
Automatic image colorization models~\cite{zhang2016cic, su2020instance, yoo2019memopainter, cao2017udc} take a gray-scale or sketch image as an input and generate a colorized image. 
CIC~\cite{zhang2016cic} proposed a fully automatic colorization model using convolutional neural networks (CNNs), and Su \textit{et al.}~\cite{su2020instance} further improved the model by extracting the features of objects in the input image. 
Despite the substantial performances of automatic colorization models, a nontrivial amount of user intervention is still required in practice.

Conditional image colorization models attempt to resolve these limitations by taking reference images~\cite{lee2020referencebased} or user interactions~\cite{zhang2017realtime, ci2018alacgan, zhang2018style2paints, xiao2018interactive, ZhangLS0WL21} as additional input. 
For example, Zhang \textit{et al.}~\cite{zhang2017realtime} allowed the users to input the point-wise color hint in real time, and AlacGAN~\cite{ci2018alacgan} utilized stroke-based user hints by extracting semantic feature maps.
Although these studies consider the results are improved by user hints, 
they generally require a large amount of user interactions.

\begin{figure*}[t!]
\centering
\includegraphics[width=\textwidth]{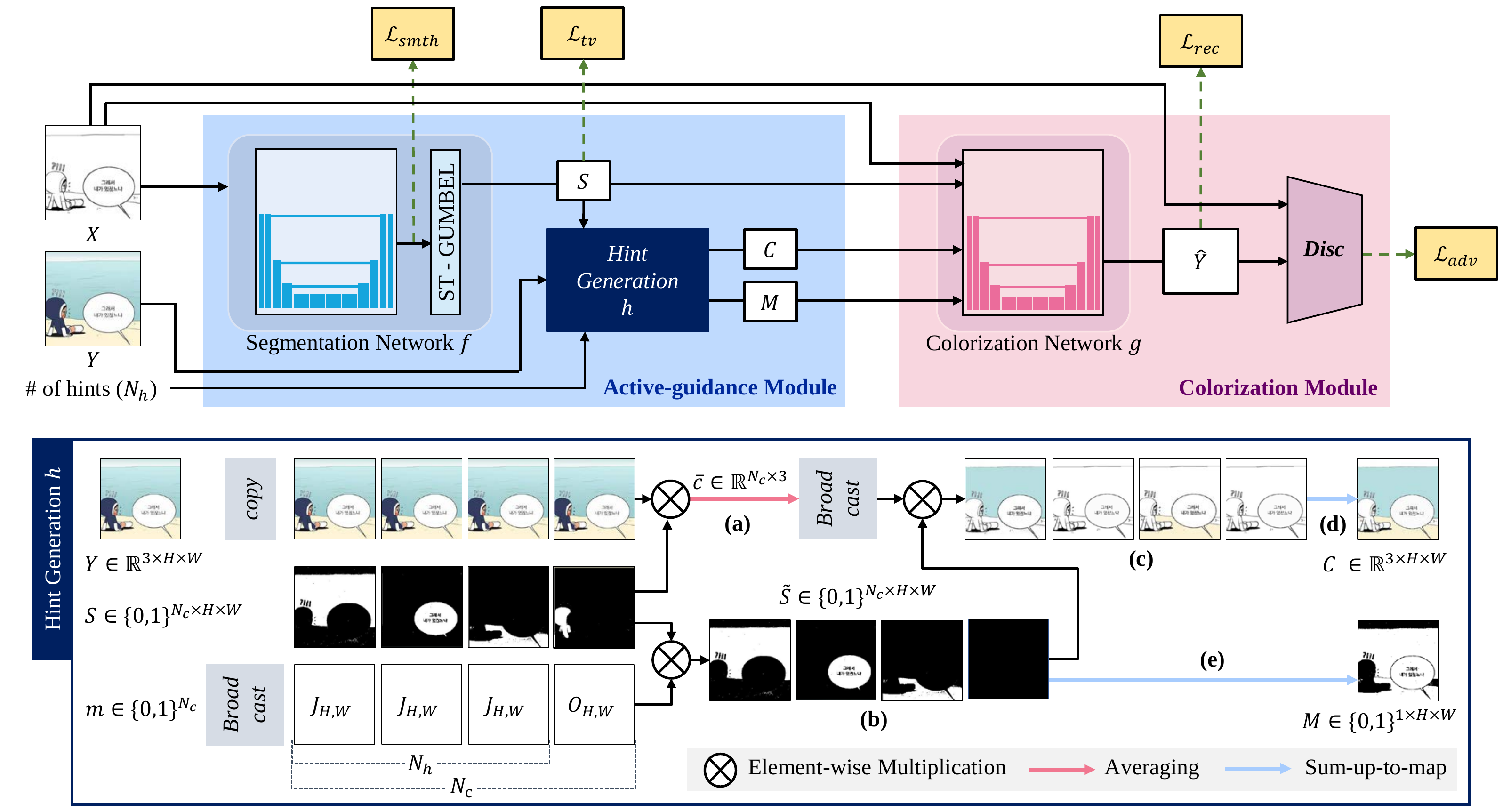}
\vspace{-0.25in}
\caption{
\textbf{Hint generation process of our proposed \model model.}
The \segnet and the \hintgen renders colored hints ($C$) and condition masks ($M$). 
Based on the guidance results, our \colornet colorizes the sketch image. 
The example illustrates the hint generation process in the training phase where $N_h=3$ and $N_c=4$. 
First, the groundtruth image is copied as $N_c$ times to consider each color segment at each interaction step. 
After element-wise multiplication with guided regions, (a) averages the color to decide representative colors for each guided region.
To restrict the number of hints, we mask out the segments whose iteration step is larger than $N_h$, The masked results are (b). 
Based on (a) and (b), our module generates the colored condition for each segment as (c). 
In (d), we combine them into one partially-colorized image $C$. (e) operates as the same manner with (d) and generates the condition mask $M$.
}
\vspace{-0.25in}
\label{fig:main_architecture}
\end{figure*}


\subsection{Interactive Image Generation} 
Beyond the colorization task, user interaction is utilized in numerous computer vision tasks, such as image generation, and image segmentation. 
In image generation, research has been actively conducted to utilize various user interactions as additional input to GANs. 
A variety of GAN models employ image-related features from users to generate user-driven images~\cite{heim2019constrained, li2020deep} and face images~\cite{portenier2018faceshop, jo2019scfegan, tang2020bipartite, lee2020maskgan, tan2020michigan}.
Several models generate and edit images via natural-language text~\cite{xu2017attngan, nam2018textadaptive, zhu2018generative, cheng2020sequential}.  
In image segmentation, to improve the details of segmentation results, recent models have utilized dots~\cite{roth2020going, majumder2020localized} and texts~\cite{hu2016segmentation} from users. 
Although we surveyed a wide scope of interactive deep learning models beyond sketch image colorization, there is no directly related work with our approach, to the best of our knowledge. Therefore, the use of a deep learning-based guidance system for interactive process can be viewed as a promising but under-explored approach. 

\section{Proposed Approach}
\subsection{Problem Setting}
\label{sec:method_problem_definition}
The goal of the interactive colorization task is to train networks to generate a colored image $\hat{Y}\in\mathbb{R}^{3\times H\times W}$ by taking as input a sketch image $X\in \mathbb{R}^{1 \times H\times W}$ along with user-provided partial hints $U$, where $H$ and $W$ indicate the height and width of the target image, respectively.
The user-provided partial hints are defined as a pair $U=(C, M)$ where $C\in\mathbb{R}^{3\times H \times W}$ is a sparse tensor with RGB values, and $M \in \{0,1\}^{1 \times H \times W}$ is a binary mask indicating the region in which the color hints are provided. 
Our training framework consists of two networks and one function: 
\segnet $\mathbf{f}$~(Section~\ref{sec:method_segment-guidence}), \colornet $\mathbf{g}$ (Section~\ref{sec:method_colorization_networks}), and  a \hintgen called $\mathbf{h}$ (Section~\ref{sec:method_hint_generation}), which are trained in an end-to-end manner.

\subsection{Segmentation Network}
\label{sec:method_segment-guidence}
The purpose of \segnet $\mathbf{f}(\cdot)$ is to divide the sketch input $X$ into several semantic regions which are expected to be painted in a single color, i.e., 
\begin{equation}
    S = \mathbf{f}(X; \theta_f),
\end{equation}
where $S=(S_1, S_2, ..., S_{N_c})\in \{0,1\}^{N_c\times H\times W}$, $S_i$ is the $i$-th guided region, and $N_c$ denotes the maximum number of hints.
Specifically, $\mathbf{f}$ contains an encoder-decoder network with skip connections, based on U-Net~\cite{isola2017pix2pix} architecture, to preserve the spatial details of given objects.

Since each guided region will be painted with a single color, we have to segment the output of U-Net in a discrete form while taking advantages of end-to-end learning. 
To this end, after obtaining an output tensor $S_{logit}\in \mathbb{R}^{N_c\times H \times W}$ of U-Net, we discretize $S_{logit}$ by applying straight-through (ST) gumbel estimator~\cite{Eric2017JangGP, Chris2017MaddisonMT} across channel dimensions to obtain $S$ as a differentiable approximation.
The result $S$ satisfies $\sum_{i=1}^{N_c} S_i(j)=1$ where $S_i(j)$ indicates the $i$-th scalar value of the $j$-th position vector, i.e., every pixel is contained in only one guided region. Here, $S_i(j)=1$ indicates that the $j$-th pixel is contained in the $i$-th guided region while $S_i(j)=0$ indicates that the pixel is not contained in the guided region.

\subsection{Hint Generation}
\label{sec:method_hint_generation}
The \hintgen $\mathbf{h}(\cdot)$ is a non-parametric function that plays the role of simulating $U$ based on $S$, a colored image $Y$, and the number of hints $N_h$, i.e., 
\begin{equation}
    U = \mathbf{h}(S, Y, N_h).
\end{equation}
To this end, we first randomly sample $N_h$ from a bounded distribution which is similar to a geometric distribution formulated as
\begin{equation}
\label{eq:bounded_geometric}
    \mathcal{G}(N_h=i)=
    \begin{cases}
    (1-p)^i p &\mbox{if } i=0,1,...,N_c-1 \\
    (1-p)^{N_c} & \mbox{if } i = N_c, 
    \end{cases}
\end{equation}
where $p<1$ is a hyperparameter indicating the probability that the user stops adding a hint on each trial. 
We set $N_c=30$ and $p=0.125$ for the following experiments.

\noindent \textbf{Step1: building masked segments $\tilde{S}$.}
Given $N_h$, we construct a mask vector $m \in \{0, 1\}^{N_c}$ having each element with the following rule:
\begin{equation}
\label{eq:m_i}
    m_i = 
    \begin{cases}
        1 & \mbox{if }i \leq N_h \\
        0 & \mbox{otherwise}, 
    \end{cases}
\end{equation}
where $m_i$ indicates the $i$-th scalar value of the vector $m$. 
Afterwards, we obtain a masked segment $\tilde{S} \in \mathbb{R}^{N_c \times H\times W}$ by element-wise multiplying the $i$-th element of $m$ with the $i$-th channel of $S$ as
\begin{equation}
\label{eq:masked_segments}
    \tilde{S}_i = m_i S_i, 
\end{equation}
where $S_i$, $\tilde{S}_i \in \mathbb{R}^{1\times H\times W}$ denote the $i$-th channel of $S$ and $\tilde{S}$, respectively. 

\noindent \textbf{Step2: building hint maps $C$.}
The goal of this step is to find the representative color value of the activated region in each segment $\tilde{S}_i$, and then to fill the corresponding region with this color.
To this end, we calculate a mean RGB color $\bar{c_i} \in \mathbb{R}^3$ as
\begin{equation}
\label{eq:mean_color}
    \bar{c_i} = 
    \begin{cases}
         \frac{1}{N_p}\sum_{j}^{HW}{S_i(j) \odot Y(j)} & \mbox{if }1 \leq N_p \\
        0 & \mbox{otherwise}, 
    \end{cases}
\end{equation}
where $N_p=\sum_{j}{S_i(j)}$ indicates the number of activated pixels of the $i$-th segment, $\odot$ denotes an element-wise multiplication, i.e., the Hadamard product, after each element of $S_i$ is broadcast to the RGB channels of $Y$, and both $S_i(j)$ and $Y(j)$ indicate the $j$-th position vector of each map. 
Finally, we obtain hint maps $C \in \mathbb{R}^{3 \times H\times W}$ as
\begin{equation}
\label{eq:color_map}
    C = \sum_{i=1}^{N_c}{\bar{c_i} \tilde{S}_i}, 
\end{equation}
where $\bar{c_i}$ is repeated to the spatial axis as the form of $\tilde{S}_i \in \mathbb{R}^{1\times H\times W}$ similar to Eq.~\eqref{eq:masked_segments} and $\tilde{S}_i$ is broadcast to the channel axis as the form of $\bar{c_i} \in \mathbb{R}^{3}$ as in  Eq.~\eqref{eq:mean_color}.
In order to indicate the region of given hints, 
we simply obtain a condition mask $M \in \mathbb{R}^{1\times H\times W}$ as
\begin{equation}
\label{eq:mask_map}
    M = \sum_{i=1}^{N_c}\tilde{S}_i.
\end{equation}
Eventually, the output of this module $U=C \oplus M \in \mathbb{R}^{4\times H\times W}$ where $\oplus$ indicates a channel-wise concatenation. Fig.~\ref{fig:main_architecture} illustrates overall  scheme of the hint generation process. At the inference time, we can create $U$ similar to the hint generation process, but without an explicit groundtruth image. Note that a sketch image is all we need to produce $\tilde{S}$ at the inference time. We can obtain $C$ and $M$ by assigning a color to each $S_i$ for $i=1,2,..., N_h$.

To understand how the hint generation module works, 
recall that $N_h$ is randomly sampled from the bounded geometric distribution $\mathcal{G}$ (Eq.~\eqref{eq:bounded_geometric}) per mini-batch at the training time. 
Since the probability that $i \leq N_h$ is higher than the probability that $j \leq N_h$ for $i<j$, $S_{i}$ is more frequently activated than $S_{j}$ during training the model. 
Hence, we can expect the following effects via this module:
i) $N_h$ affects in determining how many segments starting from the first channel of $S$ as computed in Eq.~(\ref{eq:m_i}-\ref{eq:masked_segments}); therefore, this mechanism encourages the \segnet $\mathbf{f}(\cdot)$ to locate relatively important and uncertain regions at the forward indexes of $S$. 
Section~\ref{sec:exp_segment-guide_module} shows this module behaves as our expectation.
ii) We can provide more abundant information for the following colorization networks $\mathbf{g}(\cdot)$ than previous approaches \textit{without requiring additional labels at training time or even interactions at test time}, helping to generate better results even with fewer hints than baselines (Section~\ref{sec:exp_comparison}). 

\subsection{Colorization Network}
\label{sec:method_colorization_networks}
The colorization network $\mathbf{g}(\cdot)$ aims to generate a colored image $\hat{Y}$ by taking all the information obtained from the previous steps, i.e., a sketch image $X$, guided regions $S$, and partial hints $U$, as
\begin{equation}
    \hat{Y} = \mathbf{g}(X, S, U; \theta_g).
\end{equation}
The reason for using the segments as input is to provide information about the color relationship, which the \segnet infers. 
In order to capture the context of the input and to preserve the spatial information of the sketch image, our colorization networks also adopt the U-Net architecture, the same as in the \segnet. 
We then apply a hyperbolic tangent activation function to normalize the output tensor of the U-Net.

\subsection{Objective Functions}
\label{sec:method_objective_functions}
As shown in Fig.~\ref{fig:main_architecture}, our networks are trained using the combined following objective functions.
For simplicity, $\mathbf{G}$ denotes the generator of our approach which contains all the procedures, i.e., $\mathbf{f}, \mathbf{h}, \mathbf{g}$, mentioned above while $\mathcal{D}$ denotes training datasets.

\noindent \textbf{Smoothness loss.} 
Although adjacent pixels in an image have similar RGB values, our segment guidance networks do not have an explicit mechanism to generate segments containing those locally continuous pixels. 
To improve the users' ability to interpret the segments, we introduce smoothness loss, as
\begin{equation}
\label{eq:smth_loss}
\mathcal{L}_{smth}=\mathbb{E}\left[ \sum_i^{HW}\sum_{j\in N_i}||S_{logits}(i)-S_{logits}(j)||_1 \right], 
\end{equation}
where $N_i$ denotes a set of eight nearest neighbor pixels adjacent to the $i$-th pixel, and $S_{logit}(i)$ indicates the $i$-th position vector of $S_{logit}$.

\noindent \textbf{Total variance loss.} 
In our framework, the quality of segments from $\mathbf{f}$ is important because the hints $U$ are built based on guided regions $S=\mathbf{f}(X)$.
Although the $f$ can be indirectly trained by the colorization signal, we introduce a total variance loss in order to facilitate this objective directly, i.e., 
\begin{equation}
\label{eq:tv_loss}
    \mathcal{L}_{tv} = \mathbb{E}_{X,Y\sim\mathcal{D}} 
    \left[ \sum_{i=1}^{N_c}{||(Y-\bar{c}_i)\odot S_i||^2_F} \right],
\end{equation}
where $||\cdot||_F$ denotes a Frobenius norm.
That is, $\mathcal{L}_{tv}$ attempts to minimize the color variance across pixels in each segment, which helps pixels of similar color form into the same segment.

\noindent \textbf{Reconstruction loss.}
Since both a sketch image $X$ and its corresponding partial hint $U$ are built from a groundtruth image $Y$ in the training phase, we can directly supervise our networks $\mathbf{G}$ so that it can generate an output image close to the groundtruth $Y$.
Following the previous work, we select the $\mathbb{L}_1$ distance function as our reconstruction loss, i.e., 
\begin{equation}
\label{eq:rec_loss}
    \mathcal{L}_{rec} = \mathbb{E}_{X,Y\sim\mathcal{D}, N_h\sim \mathcal{G}} \left[||\mathbf{G}(X,N_h,Y)-Y||_1 \right].
\end{equation}

\noindent \textbf{Adversarial loss.} As shown in the image generation work, we adopt an adversarial training~\cite{goodfellow2014gan} strategy, in which our generator $\mathbf{G}$ produces a natural output image enough to fool a discriminator $D$, while $D$ attempts to classify whether the image is real or fake. 
During the image colorization task, the original contents of a sketch input should be preserved as much as possible. 
Therefore, we leverage the conditional adversarial~\cite{mirza2014cgan} loss, written as
\begin{equation}
\label{eq:adv_loss}
    \begin{split}
    & \mathcal{L}_{adv} = \mathbb{E}_{X,Y\sim\mathcal{D}} \left[\log \mathbf{D}(Y, X) \right] \\ 
    & + \mathbb{E}_{X,Y\sim\mathcal{D}, N_h\sim \mathcal{G}} \left[\log (1 - \mathbf{D}(\mathbf{G}(X,N_h,Y), X)) \right].
    \end{split}
\end{equation}
Finally, our objective function is defined as
\begin{equation}
    \begin{split}
    \min_{\mathbf{G}}\max_{\mathbf{D}}\mathcal{L}=\lambda_{tv}\mathcal{L}_{tv}+\lambda_{smth}\mathcal{L}_{smth} \\ 
    +\lambda_{rec}\mathcal{L}_{rec}+\lambda_{adv}\mathcal{L}_{adv}, 
    \end{split}
\end{equation}
where each $\lambda$ indicates the weighting factor for each loss term. 
We describe the implementation details in the supplementary material.

\begin{table*}[h]
\centering
    \begin{tabular}{ l|| c | c  c  c | c  c  c | c c c }
    \hline
        ~ & ~ &\multicolumn{3}{c|}{$\mbox{PSNR}_{\uparrow}$} & \multicolumn{3}{c|}{$\mbox{FID}_{\downarrow}$} & \multicolumn{3}{c}{$\mbox{NRI}_{\downarrow}$}\\
    
      Methods                           & Cond.     & Yumi  & Tag2pix   & CelebA    & Yumi  & Tag2pix   & CelebA & Yumi  & Tag2pix   & CelebA \\
      \hline
      \hline
      CIC           & \xmark    & 15.17 & 13.99     & \underline{17.01}     & 137.35   & 167.88 & 79.05 & \multicolumn{3}{c}{-}\\
      Pix2Pix   & \xmark    & 15.11 & \underline{14.68} & 16.22 & 71.93 & 111.45 & 54.86 & \multicolumn{3}{c}{-}  \\
      AlacGAN      & \xmark    & 15.02 & 14.12 & 15.73 & \textbf{30.72} & \textbf{46.24} & \textbf{23.52} & \multicolumn{3}{c}{-} \\
      RTUG     & \xmark    & \textbf{19.05} & 14.44 & \textbf{17.23} & 35.52 & 92.69 & 52.67 & \multicolumn{3}{c}{-}\\
      \hline
      Ours                              & \xmark    & \underline{18.63} & \textbf{15.19} & 16.53 & \underline{34.07} & \underline{55.31} &  \underline{42.46} & \multicolumn{3}{c}{-} \\
      
      \hline
      \hline
      
      AlacGAN      & \cmark    & 15.68 & 14.53 & 16.52 & \underline{29.74} & \underline{46.52}  & \underline{22.83} & 31.00 & 31.00 & 31.00 \\
      RTUG     & \cmark    & \underline{20.10} & \underline{16.36} & \underline{19.16} & 30.26 & 63.58 & 44.45 & \underline{13.82} & \underline{14.79} & \underline{11.64} \\
      \hline
      Ours                              & \cmark    & \textbf{20.88} & \textbf{17.55} & \textbf{20.24} & \textbf{24.46} & \textbf{43.18} & \textbf{16.43} & \textbf{11.08} & \textbf{11.39} & \textbf{6.98} \\
      \hline
    \end{tabular}

  \caption{
  \textbf{Quantitative comparisons} in terms of PSNR, FID, and NRI (Section~\ref{sec:exp_eval_metrics}). For conditional cases, we compute the expected values of PSNR and FID when the number of synthesized hints follows $\mathcal{G}$.
  }
  \label{tab:quanti_baseline_main}
  \vspace{-0.1in}
\end{table*}

\section{Experiments}
\subsection{Sketch Image Datasets}
\label{sec:exp_datasets}

\noindent \textbf{Yumi's Cells}~\cite{yumicells} is composed of 10K images from 509 episodes of a web cartoon, named \textit{Yumi's Cells}, where a small number of characters appear repeatedly. 
Because it was published in a commercial industry, this dataset includes not only character objects but also non-character objects, e.g., text bubbles, letters, and background gradation.
Therefore, we chose this dataset to evaluate the practical effectiveness of our model.

\noindent \textbf{Tag2pix}~\cite{Kim2019tag2pix} consists of over 60K filtered large-scale anime illustrations from the Danbooru dataset~\cite{danbooru2017}. 
While this dataset consists of images of a single character and a simply colored background, the diversity of each character in terms of pose and scale makes it challenging to generate plausible colored outputs. 
We chose this dataset to verify that our model reflects various user hints well. 

\noindent \textbf{CelebA}~\cite{liu2015celeba} is a representative dataset which contains 203K human face images from diverse races.
We chose it to evaluate our model on real-world images rather than artificial ones. 
We randomly divided each dataset into a training, a validation, and a test set with the ratio of 81:9:10 and resize all images to $256\times 256$. Referring to the recipe of Lee \textit{et al.} (2020)~\cite{lee2020referencebased}, the sketch images were extracted using the XDoG~\cite{winnemoller2012xdog} algorithm.

\subsection{Evaluation Metrics}
\label{sec:exp_eval_metrics}
\noindent \textbf{Peak signal to noise ratio (PSNR)} has been broadly used as a pixel-level evaluation metric for measuring the distortion degree of the generated image in the colorization tasks~\cite{zhang2016cic, isola2017pix2pix}.
The metric is computed as the logarithmic quantity of the maximum possible pixel value of the image divided by the root mean squared error between a generated image and its groundtruth.

\noindent \textbf{Frechét inception distance (FID).} 
We used FID~\cite{heusel2017fid} as an evaluation metric for measuring the model performance by calculating the \textit{Wasserstein}-2 distance of feature space representations between the generated outputs and the real images.
A low FID score means that the generated image is close to the real image distribution.

\noindent \textbf{Number of required interactions (NRI).} 
We propose a new evaluation metric to measure how many user interactions are required for the model to produce an image of a certain quality. 
To this end, we count the number of hints needed by the model to reach a benchmark of PSNR.
If the model cannot reach a certain level of accuracy even with the maximum number of hints, we compute the count as the maximum number of hints plus one.
The benchmark can be set according to the user's tolerance or the purpose of a framework. 
We set 20.5, 17.5, and 19.5 as the benchmarks for Yumi, Tag2pix, and CelebA datasets, respectively.

\begin{figure}[t]
\centering
\includegraphics[width=\linewidth]{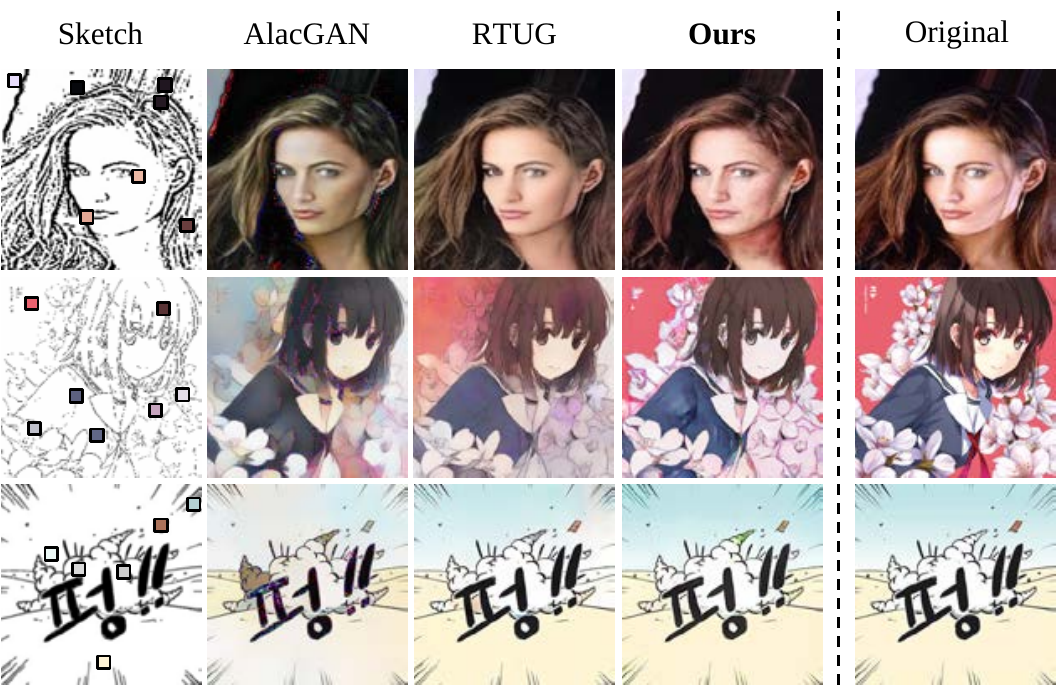}
\vspace{-0.2in}
\caption{\textbf{Comparison to baselines on diverse datasets.} 
We compare our model with two conditional colorization baselines, AlacGAN and RTUG. 
From top to bottom, conditional results on CelebA, Tag2pix, and Yumi's Cells datasets are presented.}
\label{fig:qual_comp}
\vspace{-0.1in}
\end{figure}

\subsection{Comparisons against Colorization Baselines}
\label{sec:exp_comparison}

We compare our \model against diverse baseline models for the deep colorization tasks, including an image translation model Pix2Pix~\cite{isola2017pix2pix}, an automatic colorization model CIC~\cite{zhang2016cic}, a point-based interactive colorization model RTUG~\cite{zhang2017realtime}, and a scribble-based interactive colorization model AlacGAN~\cite{ci2018alacgan}. 
Since our main focus is interactive colorization, we primarily analyze the performance of \model using conditional cases.
In order to analyze the colorization efficiency of conditional models, we compute NRI and the expected values of PSNR and FID when the number of color hints follows the distribution $\mathcal{G}$. The color hints are synthesized by their own method used for training, i.e., RTUG and AlacGAN provide hints in random location while our model provides color hints to the regions obtained by the \compseg, in order from the front channel ($S_1, S_2, ...$). 
Table~\ref{tab:quanti_baseline_main} presents the quantitative results of our model and other baselines on each dataset.
Our model outperforms all of conditional baselines on all three metrics. 
This reveals that our model can generate various realistic images, reflecting the given conditions while reducing the interactions.
Although our framework is mainly designed to colorize sketch images when conditions are given, our model shows the comparable performances across the automatic colorization setting.
We also analyze the effectiveness of our guidance mechanism in situations where real users give hints in Section~\ref{sec:user_study}.

As shown in Fig.~\ref{fig:qual_comp}, our model colorizes each color within each segment by successfully reflecting both the location and the color of hints.
The results show that ours is better than other conditional baselines. 
For a fair qualitative comparison, we equalize the number of hints given to each method and make the locations of the color hints for AlacGAN and RTUG similar to ours, by sampling the points in the regions that our segmentation network produces. 
The marks in the sketch image in Fig.~\ref{fig:qual_comp} indicate where the hints are provided for RTUG.
Compared with the conditional baselines on the animation dataset, our model reduces the color bleeding artifact, e.g., the second row in Fig.~\ref{fig:qual_comp},
and generates the continuous colors for each segment, e.g., hair in the first row, the sky and the ground in the third row in Fig.~\ref{fig:qual_comp}.
This reveals that our model can distinguish the semantic regions of character and background and reflect the color hints into the corresponding regions.
Especially, for the last two rows of Fig.~\ref{fig:qual_comp}, our model is superior to colorize the background region, while other baselines colorize the background across the edges or only part of the object. Technically, our approach can be applied to colorize not only a sketch image but also a gray-scale image.
Additional results for qualitative comparison and grayscale colorization are in the supplementary materials.

\begin{table}[t]
\centering
\begin{tabular}{c | c c c | c}
\hline
 & \multicolumn{3}{c|}{$\mbox{TPI}_{\downarrow}$ / $\mbox{QS}_{\uparrow}$} & $\mbox{CS}_{\uparrow}$ \\
 & \multicolumn{1}{c|}{Yumi's Cells} & \multicolumn{1}{c|}{Tag2pix} & CelebA & total \\ 
 \hline
RTUG & \multicolumn{1}{c|}{ 11.87 / 3.93 } & \multicolumn{1}{c|}{ 8.02 / 3.15 } & 7.82 / \textbf{3.85} & 3.14 \\
Ours & \multicolumn{1}{c|}{ \textbf{7.80} / \textbf{4.07} } & \multicolumn{1}{c|}{ \textbf{7.22} / \textbf{4.00} } & \textbf{7.13} / 3.81 & \textbf{4.07}\\
\hline
\end{tabular}
\caption{
\textbf{User study results on three different datasets.}
Time per interaction (TPI) is the average time (sec.) spent by a user before moving on to next interaction. 
Quality score (QS) is the overall quality of a colorized image.
Convenience score (CS) denotes users' convenience on the overall workflow. 
QS and CS are measured from one to five, and all scores were surveyed by users.}
\label{table:user_study_quant}
\end{table}

\begin{figure}[t]
\centering
\includegraphics[width=0.45\textwidth]{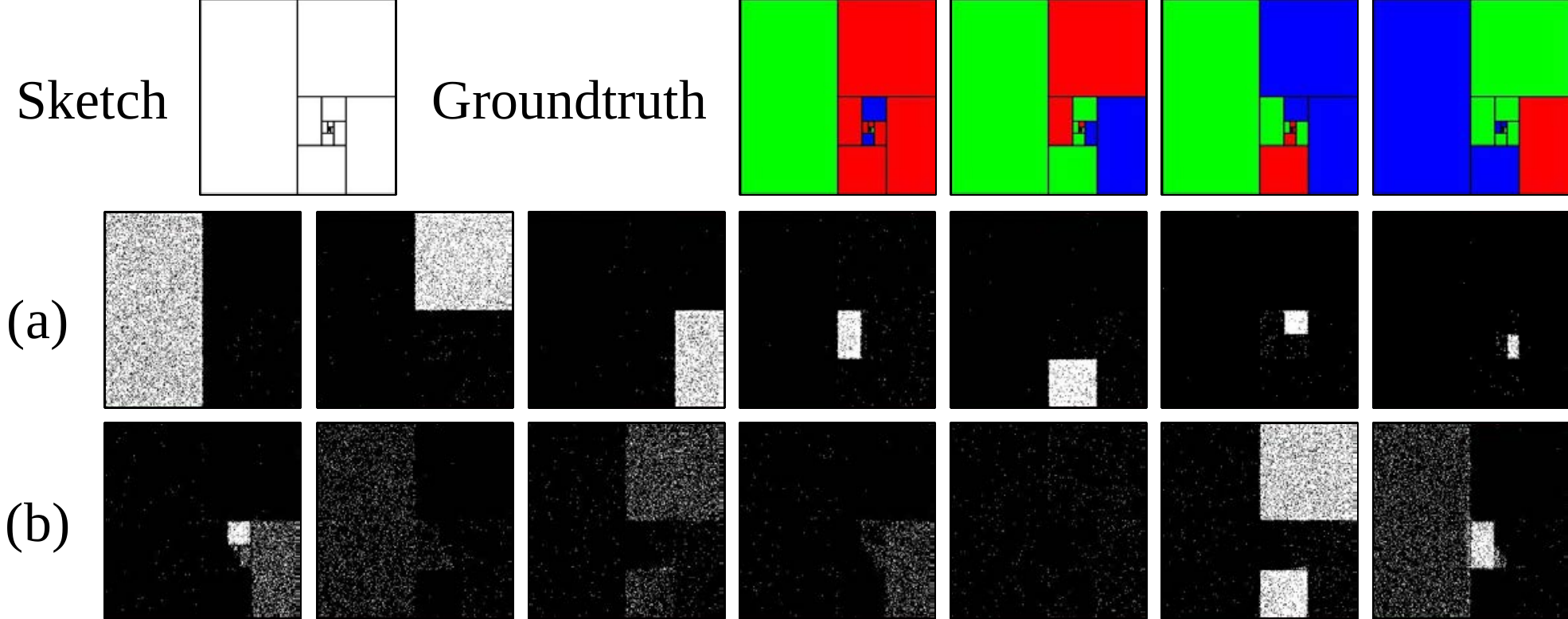}
\caption{
\textbf{Dark snail example.}
A sketch image(top left) and a groundtruth image (top right) on dark snail. 
(a) are prioritization results of \model, and (b) are when we fix $N_h=N_c$ of our model during the training time.}
\vspace{-0.1in}
\label{fig:toy_example}
\end{figure}

\subsection{User Study on Interactive Colorization Process}
\label{sec:user_study}

To validate the practical interactive process of our active-guidance mechanism, we develop a straightforward user interface (UI) that control peripheral variables except for our main algorithm. 
We conduct an in-depth user evaluation, in which users directly participate in the process of our framework. 
We then record various metrics to assess the practical usefulness of our method. 
We choose RTUG as our baseline interactive method since its interactive process is directly comparable to ours. 
As shown in Table~\ref{table:user_study_quant}, our model shows better time-per-interaction (TPI) scores with less qualitative degradation than RTUG model, confirming the superior \textit{time efficiency} of our model. 
The total colorization time is decreased by 14.2\% on average compared to
RTUG.
Furthermore, the improvement in the convenience score (CS) reveals that our approach clearly reduces the users' workload.
For more details, e.g., our UI design, see the supplementary material.

\subsection{Effectiveness of Active-Guidance Mechanism}
\label{sec:exp_segment-guide_module} 
To understand the effects of our active-guidance mechanism described in Section~\ref{sec:method_segment-guidence}-\ref{sec:method_hint_generation}, we design two sub-experiments as follows.

\noindent \textbf{Dark Snail.} The first one is a simulation to show that the proposed mechanism works as we expected by using the toy example named \textit{Dark Snail}. As shown in the first row of Fig.~\ref{fig:toy_example}, squares and rectangles are sequentially placed in a clockwise direction, and a groundtruth is generated at every mini-batch by having randomly sampled colors of red, green, and blue.
In this setting, it is impossible for a model to estimate the exact color of each object unless each color hint is provided. Because the size of each rectangle is halved compared to the previous one, querying the largest region first is an optimal choice in terms of the information gain.
In other words, this toy experiment is designed to confirm whether our model can (i) divide the semantic regions with the same color and (ii) ask for the color hints of objects in a descending order by their size. Fig.~\ref{fig:toy_example} (a) shows the guided regions obtained from a model that is trained by our original mechanism, \model. Surprisingly, the original model tends to build the semantic segments, which are i) bounded by only one object and ii) placed in decreasing order based on the segment's size, except for the 4-\textit{th} case.
Alternately, Fig.~\ref{fig:toy_example} (b) is retrieved from a modified version of our model that is trained by fixing $N_h=N_c$ during the training time, i.e., we simply turn off the most critical role of hint generation function. 
Fig.~\ref{fig:toy_example} (b) demonstrates that the modified model totally loses its guiding function, implying that the active-guidance mechanism plays a critical role in our framework.

\begin{figure}[t]
\centering
\includegraphics[width=0.45\textwidth]{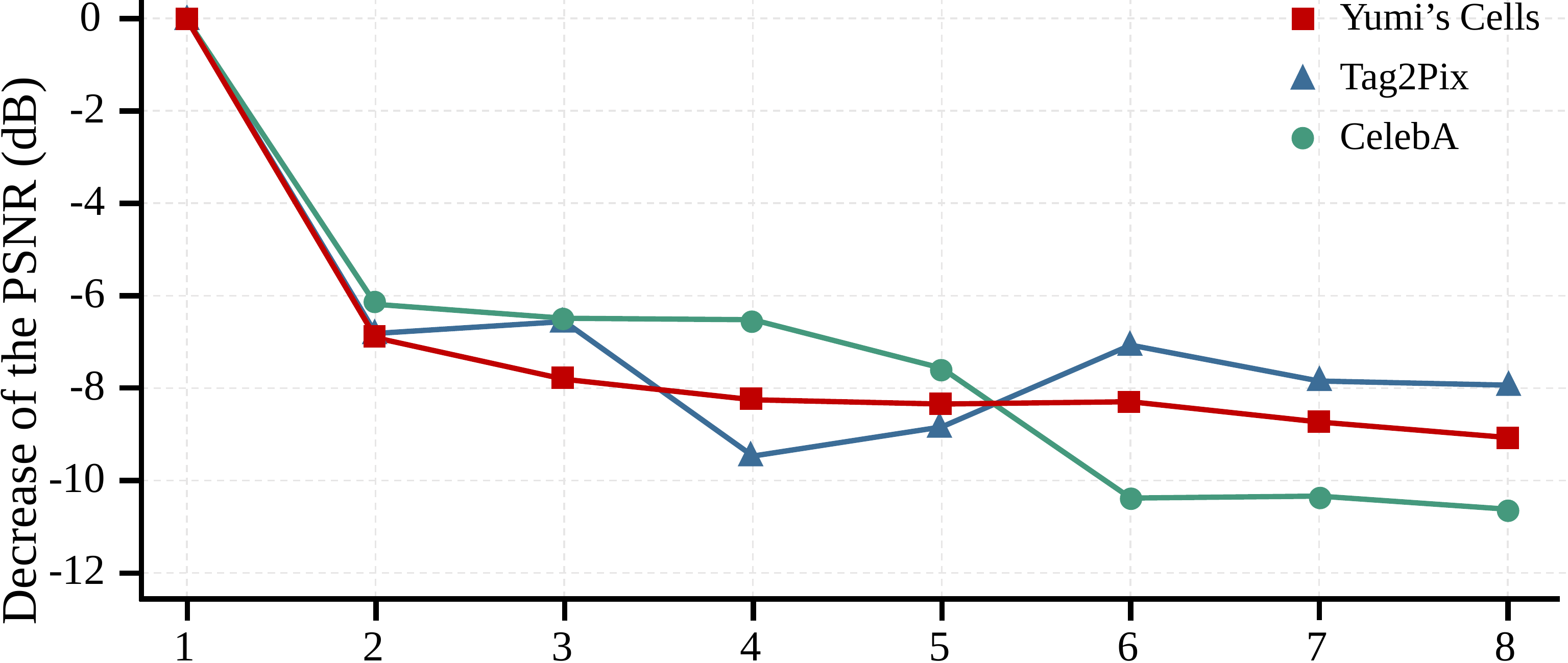}
\caption{
\textbf{PSNR scores of \model decrease} when masking out all segment information except for the $i$-th hint.}
\vspace{-0.2in}
\label{fig:segment_graph}
\end{figure}

\noindent \textbf{Importance of highly ranked segments.}
For every dataset described in Section~\ref{sec:exp_datasets}, we test how each segment provided by the active-guidance module affects the performance of colorization. To assess the importance of the $i$-th segment, we put the map of the $i$-th channel in front of remaining channels of $S$ and then give a hint only at the first segment. 
Fig.~\ref{fig:segment_graph} shows the tendency that the PSNR score decreases as a hint is given from the rear-ranked segment, which shows that the \compseg encourages to locate the important regions in the front channels of $S$.

While following the colorization order suggested by the model is an efficient way to reduce loss at training time, it is also possible to change the colorization order with additional learning. Detailed discussions on our approach, including the learning method for changing the order and limitations, are provided in the supplementary materials.

\begin{figure*}[t]
\centering
\includegraphics[width=\linewidth]{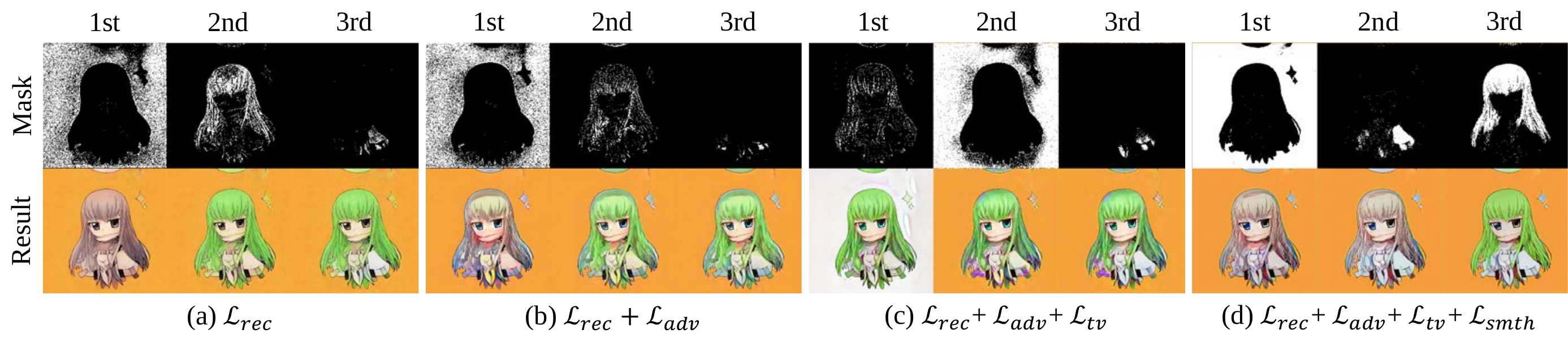}
\caption{
\textbf{Qualitative results of ablation study} for the losses (a) $\mathcal{L}_{rec}$, (b) $\mathcal{L}_{rec}+\mathcal{L}_{adv}$, (c) $\mathcal{L}_{rec}+\mathcal{L}_{adv}+\mathcal{L}_{tv}$, and (d) $\mathcal{L}_{rec}+\mathcal{L}_{adv}+\mathcal{L}_{tv}+\mathcal{L}_{smth}$. 
Each column of images indicates the priority of guided regions. The first row shows regions guided by the model in order of their priority, and the second row represents the intermediate colorization results.}
\label{fig:ablation_loss_qual}
\end{figure*}

\begin{table*}[t]
\centering
\begin{tabular}{c | c c c c | c c c}
\hline
 &\multirow{2}{*}{$\mathcal{L}_{rec}$} & \multirow{2}{*}{$\mathcal{L}_{adv}$} & \multirow{2}{*}{$\mathcal{L}_{tv}$} & \multirow{2}{*}{$\mathcal{L}_{smth}$} & \multicolumn{3}{c}{PSNR / FID / \newmetric}  \\
 & &  &  &  & \multicolumn{1}{c|}{Yumi's Cells} & \multicolumn{1}{c|}{Tag2pix} & CelebA \\ 
 \hline
(a) & \checkmark  & &  & & \multicolumn{1}{c|}{\textbf{21.04} / 28.30 / \textbf{9.95}} & \multicolumn{1}{c|}{\textbf{18.11} / 66.21 / \textbf{8.86}} & \textbf{20.76} / 39.73 / \textbf{5.58} \\
(b) & \checkmark  & \checkmark    & & & \multicolumn{1}{c|}{20.84 / 27.90 / 11.05} & \multicolumn{1}{c|}{17.55 / 48.37 / 11.30} & \underline{20.41} / 17.15 / \underline{6.13} \\
(c) & \checkmark  & \checkmark    & \checkmark & & \multicolumn{1}{c|}{20.72 / \underline{27.81} / \underline{10.39}} & \multicolumn{1}{c|}{\underline{17.58} / \underline{47.32} / \underline{10.48}} & 20.19 / \underline{16.54} / 7.20 \\
(d) & \checkmark  & \checkmark    & \checkmark & \checkmark & \multicolumn{1}{c|}{\underline{20.88} / \textbf{24.46} / 11.08} & \multicolumn{1}{c|}{17.55 / \textbf{43.18} / 11.39} & 20.24 / \textbf{16.43} / 6.98 \\
\hline
\end{tabular}
\caption{
\textbf{Quantitative results of the ablation study} for the losses (a) $\mathcal{L}_{rec}$, (b) $\mathcal{L}_{rec}+\mathcal{L}_{adv}$, (c) $\mathcal{L}_{rec}+\mathcal{L}_{adv}+\mathcal{L}_{tv}$, and (d) $\mathcal{L}_{rec}+\mathcal{L}_{adv}+\mathcal{L}_{tv}+\mathcal{L}_{smth}$.
}
\vspace{-0.1in}
\label{table:ablation_loss_quant}
\end{table*}

\subsection{Effectiveness of Loss Functions}
\label{sec:exp_loss_functions}
This section analyzes the effects of each loss function using both quantitative measurements and qualitative results.
In this ablation study, we found a trade-off between the pixel-distance-based metric, i.e., PSNR, and the feature-distribution-based metric, i.e., FID, according to the combination of loss functions.
Since $\mathcal{L}_{rec}$ exactly matches up to the PSNR, Table~\ref{table:ablation_loss_quant} (a) shows the best score of the PSNR-related measurement.
However, it does not perform well in terms of FID especially in the Tag2pix and CelebA datasets. 
This phenomenon can also be found in Fig.~\ref{fig:ablation_loss_qual} (a). The character in the first colorization result tends to be painted with grayish color, and overall colorization results loss sharpness. 
After $\mathcal{L}_{adv}$ is added, the FID scores in Table~\ref{table:ablation_loss_quant} (b) dramatically improve, along with the qualitative results in Fig.~\ref{fig:ablation_loss_qual} (b), but PSNR-based scores slightly decrease.
As discussed in a previous work~\cite{wang2003perceptual}, we guess that the PSNR score is not sufficient to measure how naturally a model can generate if only partial conditions are given. 
Although Fig.~\ref{fig:ablation_loss_qual} (b) shows plausible images, the hair in all the output images are slightly stained. 
By adding $\mathcal{L}_{tv}$, these stains are removed, and the colors become clear, as shown in Fig.~\ref{fig:ablation_loss_qual} (c). 
After adding $\mathcal{L}_{smth}$, the guided regions become significantly less sparse than before, and the strange colors on the sleeve of Fig.~\ref{fig:ablation_loss_qual} (c)'s character disappear, as shown in Fig.~\ref{fig:ablation_loss_qual} (d). 
Table~\ref{table:ablation_loss_quant} shows the FID score improves after adding $\mathcal{L}_{adv}$, $\mathcal{L}_{tv}$, and $\mathcal{L}_{smth}$ one by one from $\mathcal{L}_{rec}$ on all datasets.
Despite the trade-off, we select (d) as our total loss function, considering the qualitative improvements and the balance between the PSNR-based and FID metrics.

\section{Conclusions}
\label{sec:conclusion}
This work presents a novel interactive deep colorization framework, which enables the model to learn the priority regions of a sketch image that are most in need of color hints. 
Experimental results show that our framework improves the image quality of interactive colorization models, successfully reflecting the color hints with our active guidance mechanism.
Importantly, our work demonstrates that \model, without any manual supervision at all, can learn the ability to divide the semantic regions and rank them in decreasing order of priority by utilizing the colorization signal in an end-to-end manner. 
We expect that our approach can be used to synthesize hints for training other interactive colorization models.
Developing a sophisticated UI which integrates our region prioritization algorithm with diverse techniques, such as region refinement, remains as our future work.

\noindent\textbf{Acknowledgments} This work was supported by the Institute of Information \& communications Technology Planning \& Evaluation (IITP) grant funded by the Korean government (MSIT) (No. 2019-0-00075, Artificial Intelligence Graduate School Program (KAIST) and No. 2021-0-01778, Development of human image synthesis and discrimination technology below the perceptual threshold) and the National Research Foundation of Korea (NRF) grant funded by the Korean government (MSIT) (No. NRF-2022R1A2B5B02001913).

{\small
\bibliographystyle{ieee_fullname}
\bibliography{egbib}
}

\clearpage
\appendix
\section{Overview}

\begin{figure*}[t!]
\centering
\includegraphics[width=\linewidth]{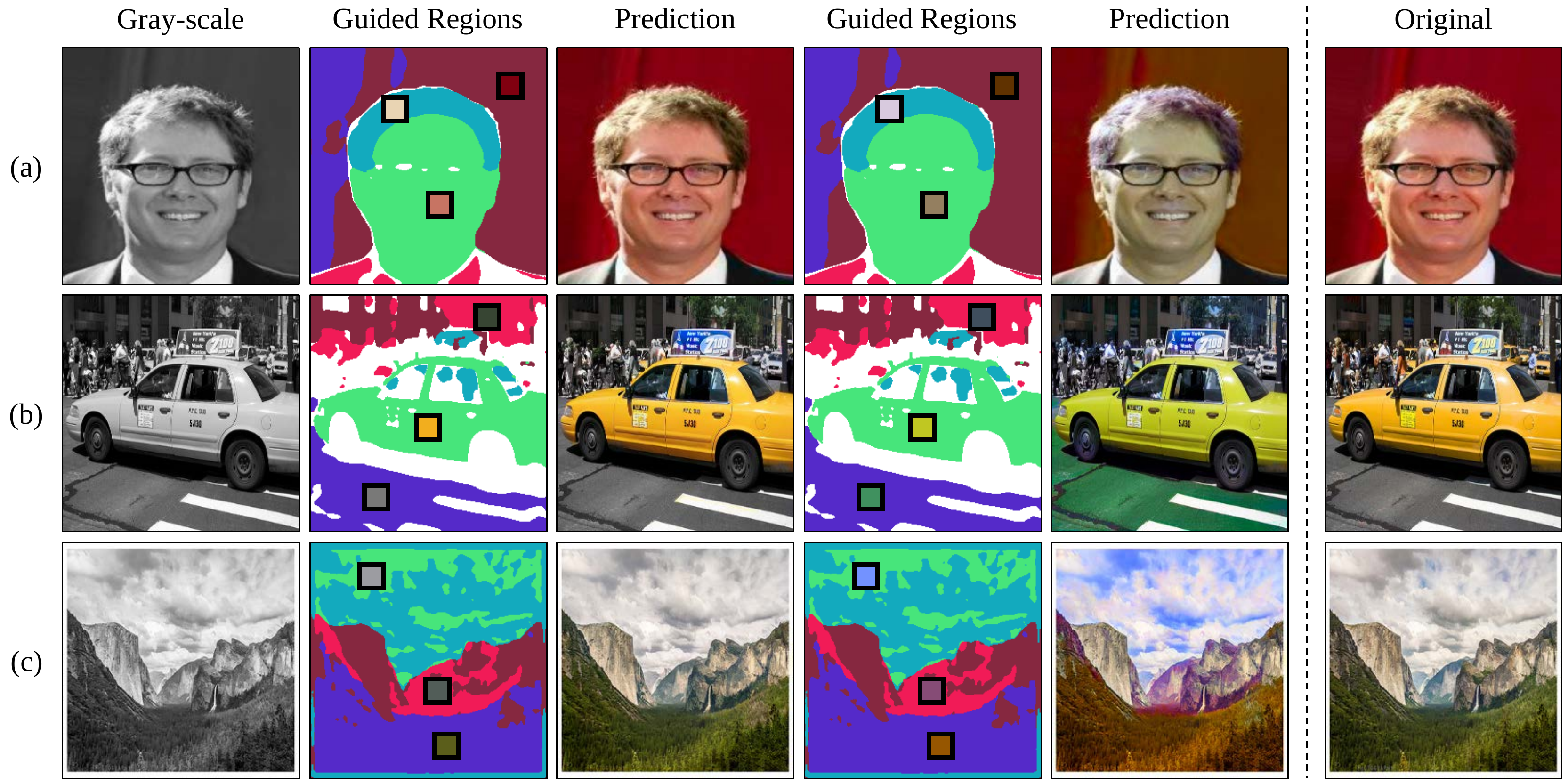}
\captionof{figure}{\textbf{Qualitative results of gray-scale colorization on (a) CelebA, (b) Imagenet Car, and (c) Summer2Winter Yosemite datasets.} First column represents gray-scale inputs. Second and fourth columns show first five guided regions generated by the segmentation network and mark regions to change the color as small squares. Third column shows the colorization results when reflecting seven original color hints and sixth column shows the colorization results when changing the color of the marked regions.}
\label{fig:grayscale}
\end{figure*}

This document addresses with additional information that we does not cover in our main paper due to the page limit. 
Section~\ref{sec:supp_grayscale_colorization} show the applicability of our approach to a gray-scale image on diverse datasets.
Section~\ref{sec:supp_implementation_details} describes the implementation details for the reproducibility, including the network architectures, the hyperparameters of optimizers we used, and the training details. 
Section~\ref{sec:supp_discussions} provides discussions on our approach, including our hinting mechanism, a method to change colorization order, and limitation of our method.
Section~\ref{sec:user_study_supp} gives detailed descriptions of the user studies we conducted in Section 4.4 of our main paper.
At the rest of this document (Section~\ref{sec:supp_qualitative_results}), we demonstrate qualitative results of our proposed method.
Note that an attached video material could help to understand the behaviors of \model in a visual manner, and source codes are provided for needs of code-level details.

\section{Application to Gray-scale Colorization}
\label{sec:supp_grayscale_colorization}

We test our approach using the gray-scale input on CelebA~\cite{liu2015celeba}, Imagenet Car~\cite{Jia2009DengDSLL}, and Summer2Winter Yosemite~\cite{Jun2017ZhuPIE} dataset. To do this, we modify the colorization model to take the L channel of an image and output the AB channel of the image.
Table~\ref{table:quant_grayscale} represents the quantitative results of RTUG and our model on each dataset.
Our model surpasses the baseline in terms of both PSNR and FID. 
This demonstrates that our model can produce realistic images while reflecting color hints in grayscale colorization. 
As shown in Fig.~\ref{fig:grayscale}, our model can guide the color hints to reasonably shaped regions and reflect the color hints.

\begin{table}[h]
\footnotesize
\begin{center}
\begin{tabular}{c | c c c}
\hline
 & \multicolumn{3}{c}{$\mbox{PSNR}_{\uparrow}$ / $\mbox{FID}_{\downarrow}$} \\
 & \multicolumn{1}{c|}{CelebA} & \multicolumn{1}{c|}{Imagenet Car} & Yosemite\\ 
 \hline
RTUG & \multicolumn{1}{c|}{ 29.92 / 3.61 } & \multicolumn{1}{c|}{ 26.40 / 31.90 } & 27.58 / 57.97 \\
Ours & \multicolumn{1}{c|}{ \textbf{30.60} / \textbf{2.40} } & \multicolumn{1}{c|}{ \textbf{26.97} / \textbf{27.54} } & \textbf{27.94} / \textbf{54.05} \\
\hline
\end{tabular}

\caption{
\textbf{Quality comparison of grayscale colorization results} in terms of PSNR, FID. We provide RTUG and our model the same amount of color hints which follows $\mathcal{G}$.}
\label{table:quant_grayscale}
\end{center}
\end{table}

\begin{table}[h]
\begin{center}
\begin{tabular}{l|l}
\hline
Label & Layer \\ \hline \hline
$E_1$ & DoubleConv($I:C_{in}$,$O:64$)  \\ \hline
$E_2$ & DoubleConv($I:64$, $O:128$) \\ \hline
$E_3$ & DoubleConv($I:128$, $O:256$) \\ \hline
$E_4$ & DoubleConv($I:256$, $O:512$) \\ \hline
$D_0$ & DoubleConv($I: 512$, $O: 512$) \\ \hline
$D_1$ & DoubleConv($I: 1024$, $O: 256$) \\ \hline
$D_2$ & DoubleConv($I: 512$, $O: 128$) \\ \hline
$D_3$ & DoubleConv($I: 256$, $O: 64$) \\ \hline
$D_4$ & DoubleConv($I: 128$, $O: 64$) \\ \hline
Out & Conv($I: 64$, $O: C_{out}$) \\ \hline
\end{tabular}
\end{center}
\caption{\textbf{The layer specification of the U-Net.} DoubleConv denotes two consecutive Conv-BatchNorm-ReLU blocks, and Conv denotes a convolution layer. $I, O$ denote the size of input channels, the size of output channels, respectively.}
\label{tab:layers}
\end{table}

\section{Implementation Details}
\label{sec:supp_implementation_details}

\noindent \textbf{U-Net Architecture.} We adopt the U-Net architecture in the segmentation network and colorization network, except for the size of channels in the input layer and output layer. 
The layer specification is shown in Table.~\ref{tab:layers}, where $(C_{in}, C_{out})$ is equal to $(1, N_c)$ for segmentation network and equal to $(N_c+5, 3)$ for colorization network. 
Maxpooling is applied to the front of each layer $E_2$-$D_0$ to downsample the input tensor by a factor of 2. 
For each $i=1...4$, the bilinear upsampled output of $D_{i-1}$ and the output of $E_{5-i}$ are concatenated in the channel dimension, and then pass through $D_i$. 
Every convolution in $E_1$-$D_4$ is applied with $3\times3$ kernel, whereas the convolution in output layer is applied with $1\times1$ kernel.

\noindent \textbf{Discriminator.}
The discriminator $D$ is implemented with PatchGAN\cite{isola2017pix2pix}, which outputs a $30\times30$ tensor. We use the LSGAN\cite{xudong2016least} objective to train the GAN architecture.

\noindent \textbf{Training Details.}
We initialize the weight of networks from the normal distribution with a mean of 0 and a standard deviation of 0.02.
The Adam optimizer~\cite{diederik2015adam} with $\beta_1=0.5, \beta_2=0.999$ is used to train our networks on all datasets.
The learning rate is fixed at 0.0002 for the first half of epochs and linearly decays to zero for the remaining half of epochs.
We schedule the temperature $\tau$ of ST gumbel estimator with exponential policy $\tau=0.1^{\text{current epoch}/\text{total epochs}}$, adopted from RelGAN~\cite{Nie2019ICLR}. 
The total numbers of epochs for Yumi's Cells, Tag2Pix, and CelebA datasets are 500, 30, and 20, respectively. 
The optimization typically takes about 1-2 days on 4 TITAN RTX GPUs.

\section{Discussions}
\label{sec:supp_discussions}

\subsection{RoI-based Hinting Mechanism}
In this study, we mainly compare our RoI-based hinting mechanism with the point and scribble-based hinting mechanisms.
Note that there are trade-offs between each method.
Point or scribble-based approaches have their own advantages in that they can utilize the location of the color hint. 
However, our goal is to colorize images with a few hints, not to colorize perfectly with plenty of time. 
We therefore focus on validating the effectiveness of our region-based guidance system in terms of interaction efficiency. 
As shown in the user study (Section~4.4 of our main paper), our region-based guidance system can reduce the average time per interaction, resulting in the improved convenience score.

Since artists spend a lot of time in adding a base color on a sketch image(s) in real-world applications, it is valid work to find efficient method to mitigate such labour-intensive process. 
In this context, our work will be able to make the labour-intensive process significantly efficient.
Although accurately and efficiently colorizing more complex images is still difficult, we expect that combining our RoI hinting mechanism with the point or scribble hinting mechanism would be one of promising works to solve this problem.

\subsection{How to Change Colorization Order}
\begin{figure}[h]
\centering
\includegraphics[width=0.45\textwidth]{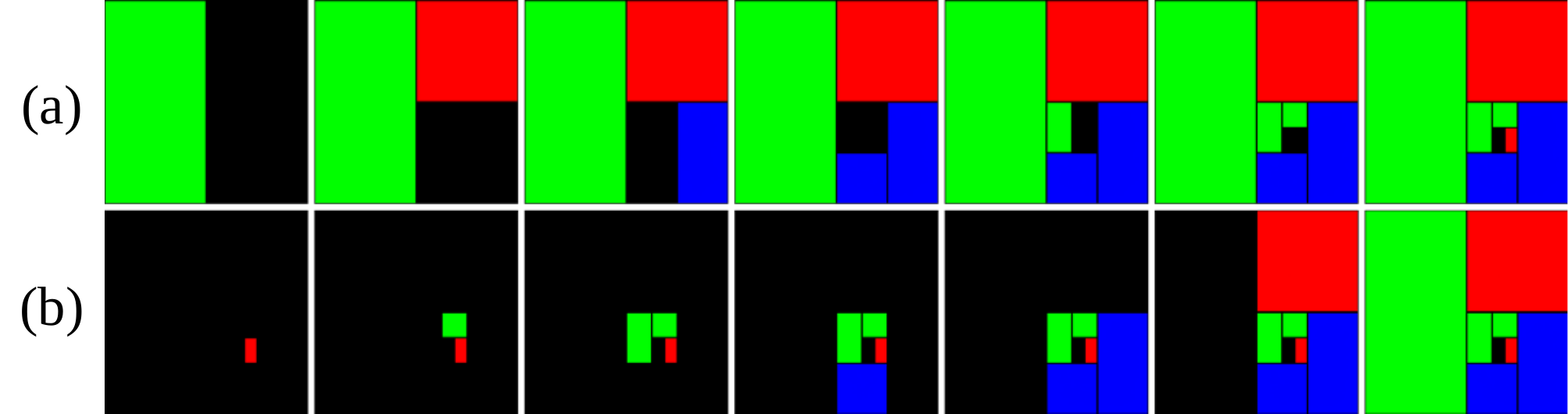}
\caption{\textbf{Colorization results of a Dark snail example in two different colorization orders.} (a) shows the colorization results when hinting regions in decreasing order of its size. Conversely, (b) shows the colorization results when hinting the small region first.}
\label{fig:toy_example_order}
\end{figure}

While \model automatically guides color hints to regions in an efficient order, the users may want to paint the regions regardless of this fixed order. We found that it is possible to change the colorization order of \model through two-stage learning. After training \model with the ordinary learning process, the segmentation networks gain the ability to estimate regions from a given sketch image. To make the colorization order changeable, we train only the colorization networks by fixing $N_h=N_c$ and randomly dropping out some hints produced by the hint generation module, i.e., letting each $m_i$ Bernoulli random variable with success probability $p=0.125$, in a second learning phase. As a result of this learning, the colorization networks can colorize the sketch image even if only some random regions are given color hints in random order. As shown in Fig.~\ref{fig:toy_example_order}, our modified \model can colorize a sketch image in different colorization orders.

\subsection{Effectiveness of the Number of Hints}
\label{sec:supp_effect_numhint}

We investigate how the performance of our model and baseline models changes as the number of hints increases. Fig.~\ref{fig:supp_effect_numhints} shows the change of PSNR and FID score when each of 2,4,6,8,10,12 hints are provided to each model. 
We found that \model mostly surpasses performances of the baselines if the same size of hints are given. 
If more than or equal to two color hints are given, our model surpasses other baseline models in both PSNR and FID scores. 
The results show that our hint guidance method enables the colorization module to effectively reflect hints.

\subsection{Limitation of Our Study}

\begin{figure}[h]
\begin{center}
\includegraphics[width=0.45\textwidth]{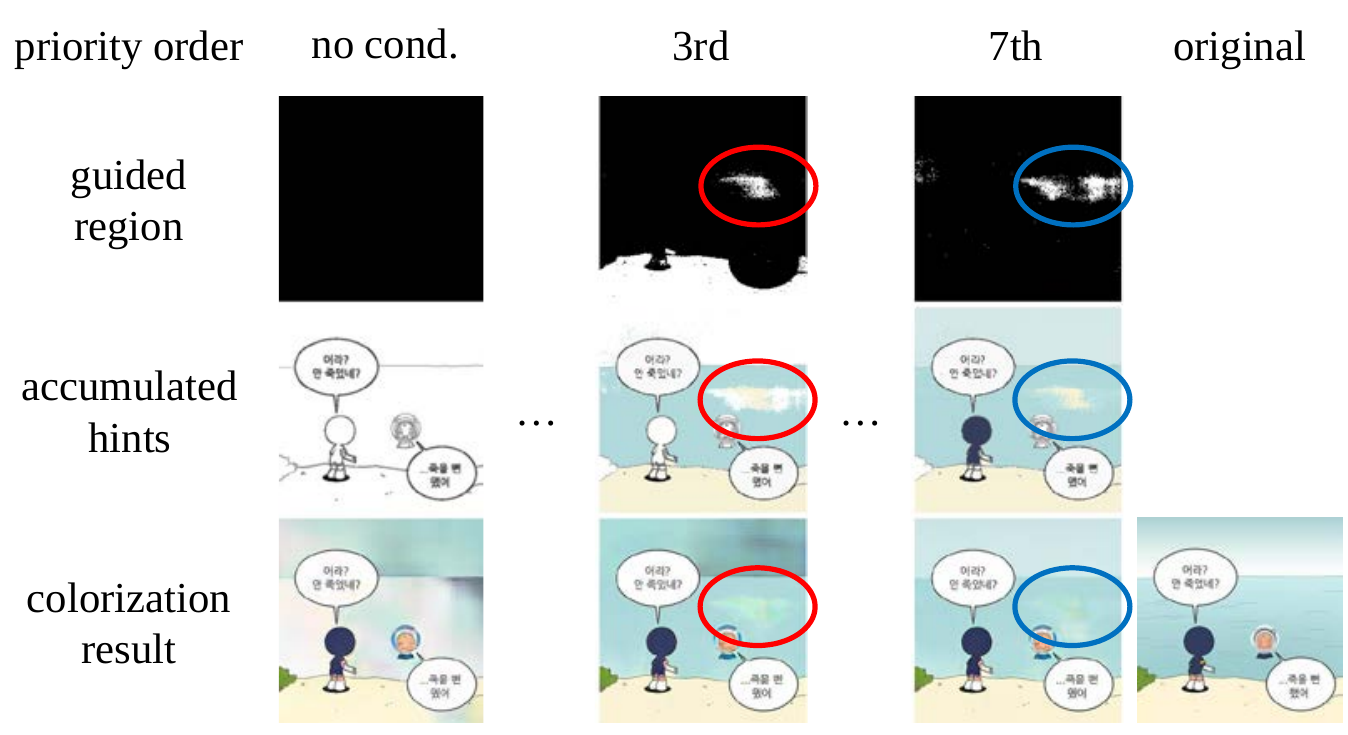}
\end{center}
\caption{\textbf{A failure case} of the segmentation network. }
\label{fig:failure_case}
\end{figure}

In some cases, the quality of the result image is degraded due to wrong prediction of the segmentation network. 
In Fig.~\ref{fig:failure_case}, the region marked as red shows a misaligned segment of the `sand' region inside the `sea' one. 
We found that these minor errors can be slightly refined by the following colorization network. 
Despite its self-correction, the stain still remains on the result, which is marked as blue.
Mitigating this problem through segmentation correction techniques would be one of promising future works.

\section{User Study}
\label{sec:user_study_supp}

This section describes details of the user study in Section 4.4 of our main paper. In addition, we conduct user-perception study to evaluate whether our model can reflect unusual color hints.

\subsection{Efficiency of Interactive Process}

\begin{figure}[h]
\begin{center}
\includegraphics[width=0.45\textwidth]{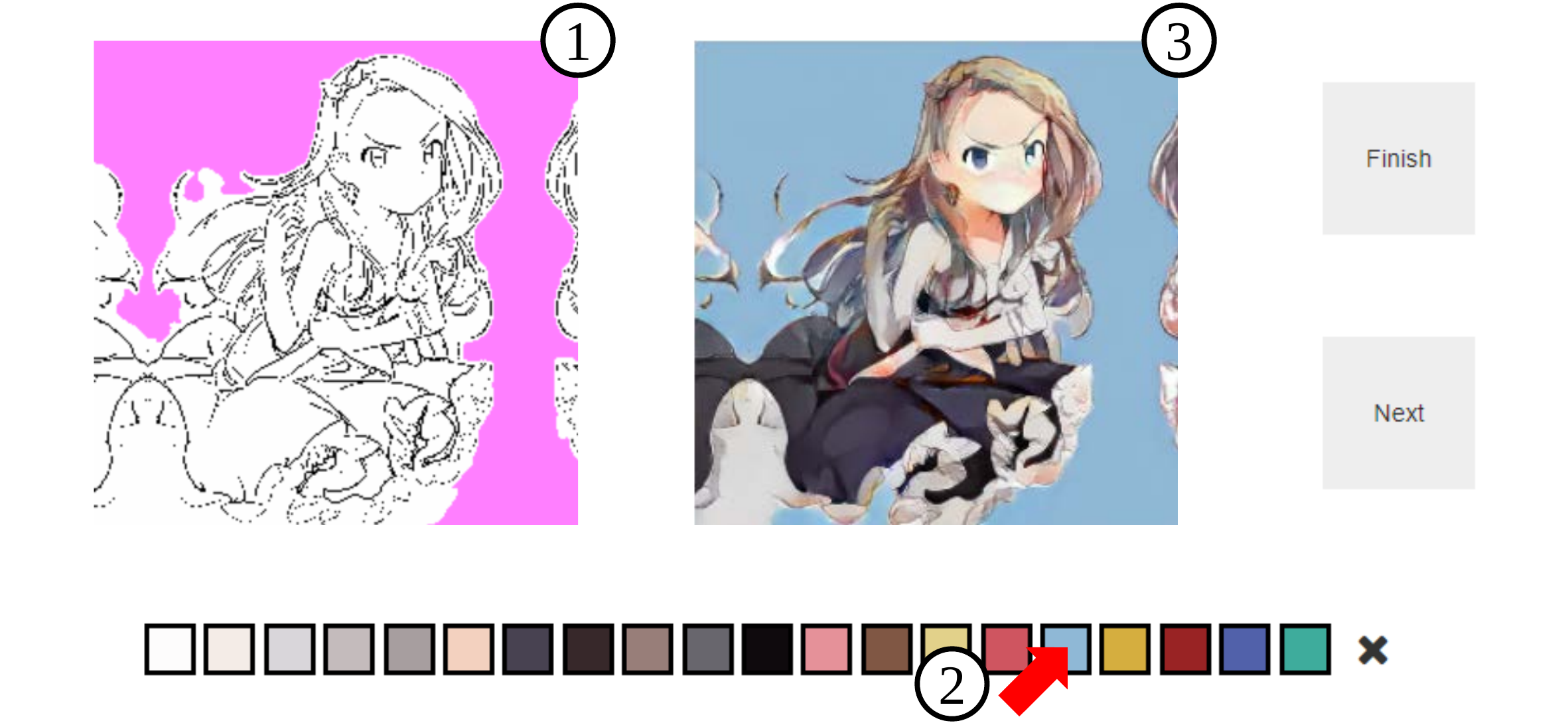}
\end{center}
\caption{\textbf{User interface of user study.} 
(1) Our model highlights recommended region to colorize.
(2) Considering the region, a user selects a color from color palettes.
(3) The result of reflecting the hint is displayed.
}
\label{fig:user_interface}
\end{figure}

Since our framework aims to increase the interaction-efficiency on colorization process, we conduct a user study to estimate how our model can enhance the overall process when the users intervene in it.
As a competitive approach to our framework for estimating interaction-efficiency, we choose RTUG since the interaction process presented in this original paper~\cite{zhang2017realtime} is most similar to ours and we can easily quantify the amount of interactions in the process.

As shown in Fig.~\ref{fig:user_interface}, we develop a straightforward user interface(UI) to rule out peripheral variables except for the main algorithms as much as possible.
The UI consists of a color palette, screens for checking hints and colorization results, and a few buttons to reflect hints or to select next hints.
The users test our method and RTUG on three datasets, Yumi's Cells, Tag2pix, and CelebA.
If the user tests our method, the user can see a region guided by our mechanism in the left image (Fig.~\ref{fig:user_interface} (1)), and choose a color from the palette at the bottom of the screen (Fig.~\ref{fig:user_interface} (2)). After selecting the color, the user can see an inference image on the right of the screen (Fig.~\ref{fig:user_interface} (3)).
This overall process is same as when the user tests RTUG, except the facts that the location of the hint is displayed in the form of a point on the left image and the user can move the point to the location by clicking the left image.
The user can click the next button to add another color hint or click the finish button to end up the colorization process.

Before the evaluation of participants, we let the users freely use the UI for about 5 minutes so that the users can be familiar with it.
A total of 13 participants comprised of researchers or engineers related to computer science and AI attend our user study. Each user is asked colorizing the given sketch image as naturally as possible without a reference image. For each dataset and method, the user completes three images using the UI.
We guide the users to finish each colorization task in roughly one minute, preventing the users from spending too much time on a single task. 
The evaluation results of the user study are shown in table 2 of our main paper.

\begin{figure}[t]
\centering
\includegraphics[width=0.45\textwidth]{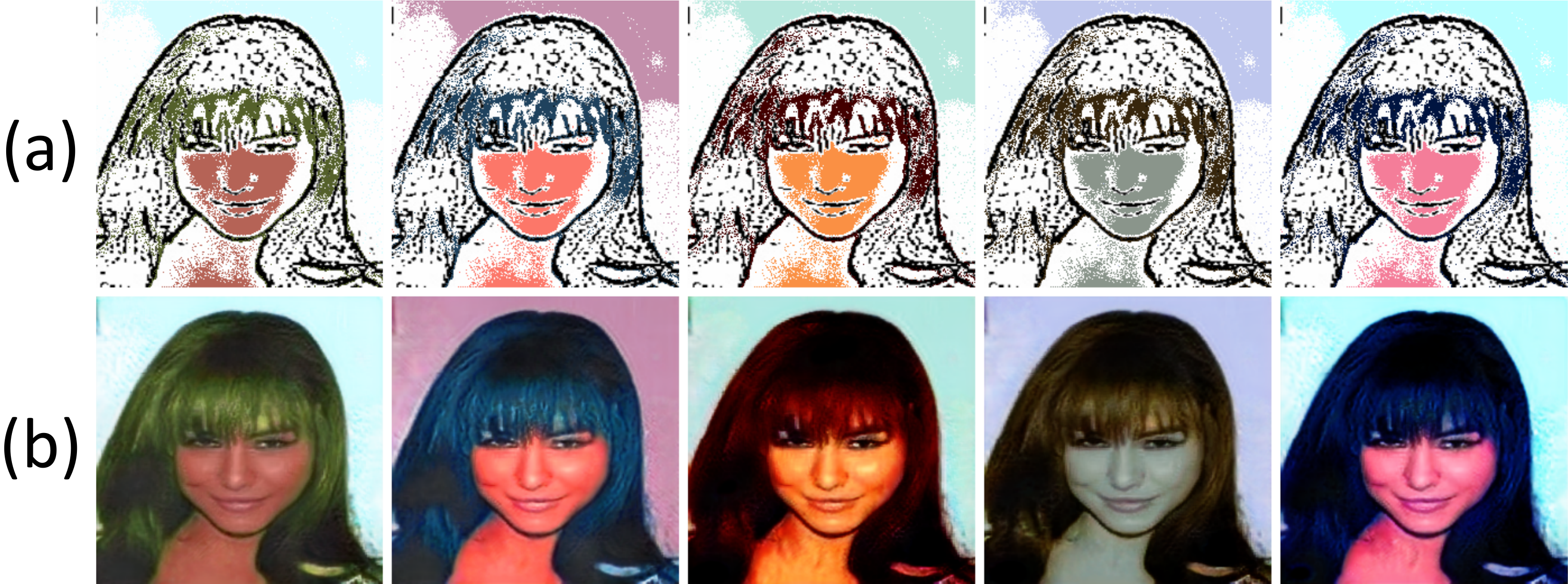}
\caption{
\textbf{Results of reflecting diverse color hints.} 
(a) represents accumulated hint images when giving diverse color hints to face, hair, and background. 
(b) shows the output images of our model for each hint image.}
\label{fig:diverse_colors}
\end{figure}

\begin{table}[t]
\begin{center}
\begin{tabular}{c | c c c}
\hline
 & \multicolumn{3}{c}{$\mbox{Top-1}_{\uparrow}$ / $\mbox{Top-2}_{\uparrow}$} \\
 & \multicolumn{1}{c|}{Yumi's Cells} & \multicolumn{1}{c|}{Tag2pix} & CelebA\\
 \hline
AlacGAN & \multicolumn{1}{c|}{ 2.7 / 12.7 } & \multicolumn{1}{c|}{ 0.0 / 10.0 } & 1.4 / 7.3 \\
RTUG & \multicolumn{1}{c|}{ 20.0 / 90.0 } & \multicolumn{1}{c|}{ 13.6 / 91.8 } & 29.1 / \textbf{97.7} \\
Ours & \multicolumn{1}{c|}{ \textbf{77.3} / \textbf{97.3} } & \multicolumn{1}{c|}{ \textbf{86.4} / \textbf{98.2} } & \textbf{69.5} / 95.0 \\
\hline
\end{tabular}
\caption{
\textbf{User-perception study results} to evaluate how faithfully models reflect user-provided conditions. 
`Top-k' indicates how many the generated images are ranked within the top-k among models over three datasets. 
All numbers are in percentages.}
\label{tab:user_study}
\end{center}
\end{table}

\subsection{Reflecting Unusual Color Hints}

We also conduct a user study to evaluate how faithfully our model and the baselines reflect the user interaction, even though the hints can contain unusual colors. 
To be specific, we randomly select 500 images for each test dataset and prepare images generated by each model with strongly perturbed color hints as shown in Fig.~\ref{fig:diverse_colors} (a). 
The perturbed color is created by adding random values between -64 and 64 to each RGB value of the groundtruth color.

For a fair comparison, we unify the locations of given hints to each model by randomly sampling them within the guided regions produced by \segnet. 
Simply, the number of provided hints and their positions are similar across the baselines and ours.
The hint images have up to seven hints, and each model generates images based on the same number of hints. 
With the generated image and the hint map, the user is asked to rank the generated images in the order of how much the hints are properly reflected. 
Table~\ref{tab:user_study} shows the percentage of generated images within the top-$k$ of the rank over all datasets. 
Our model does not only get the highest top-1 ratio over all datasets, but also successfully reflect the diverse color conditions, as shown in Fig.~\ref{fig:diverse_colors}.
This implies that our model can work robustly in terms of color variations.

\section{Qualitative Results}
\label{sec:supp_qualitative_results}

\begin{figure*}[p]
\centering
\includegraphics[width=\linewidth]{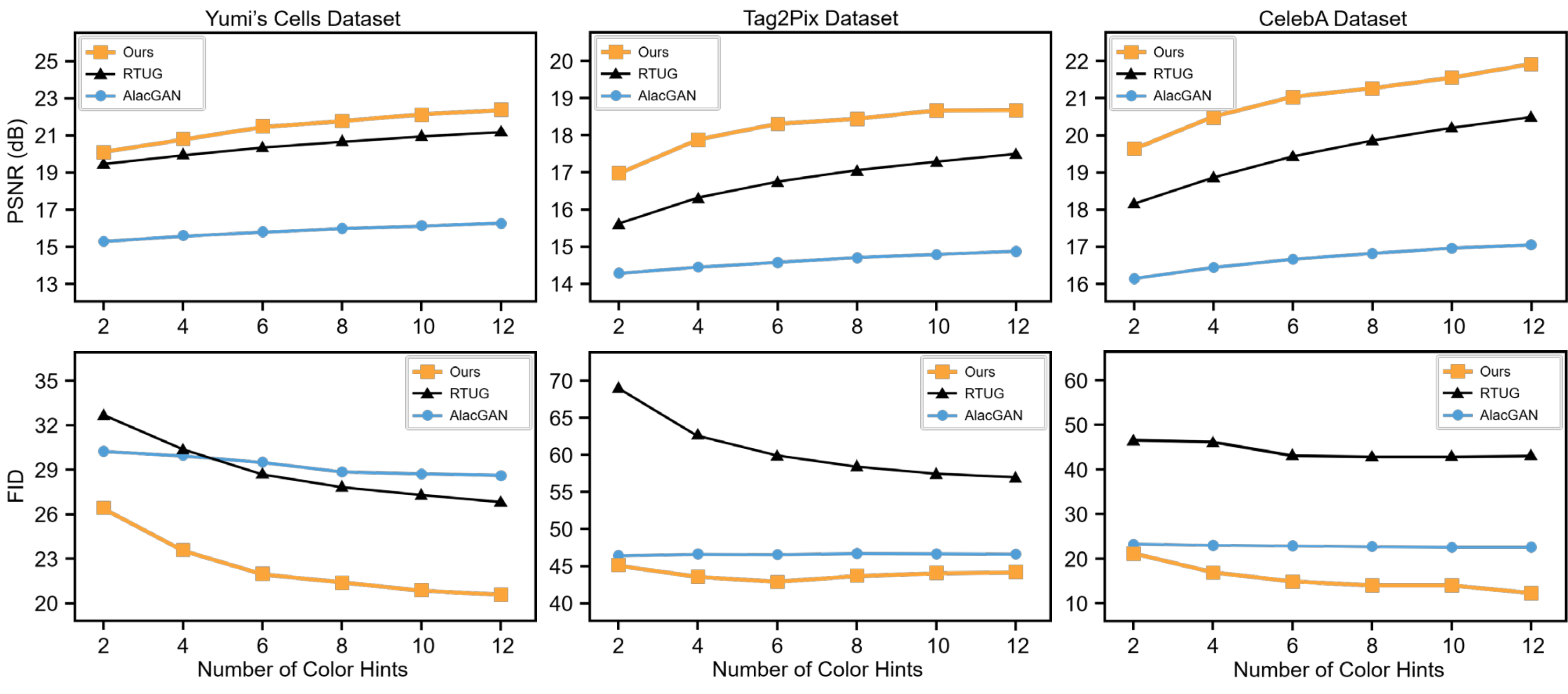}
\caption{\textbf{Performance changes according to the number of user hints.} Columns mean the results of Yumi's Cells~\cite{yumicells}, Tag2pix~\cite{Kim2019tag2pix} and CelebA~\cite{liu2015celeba} over baselines and \model. The first row shows the PSNR scores and second row presents its corresponding FID~\cite{heusel2017fid} scores. }
\label{fig:supp_effect_numhints}

\vspace{0.3in}

\centering
\includegraphics[width=\textwidth]{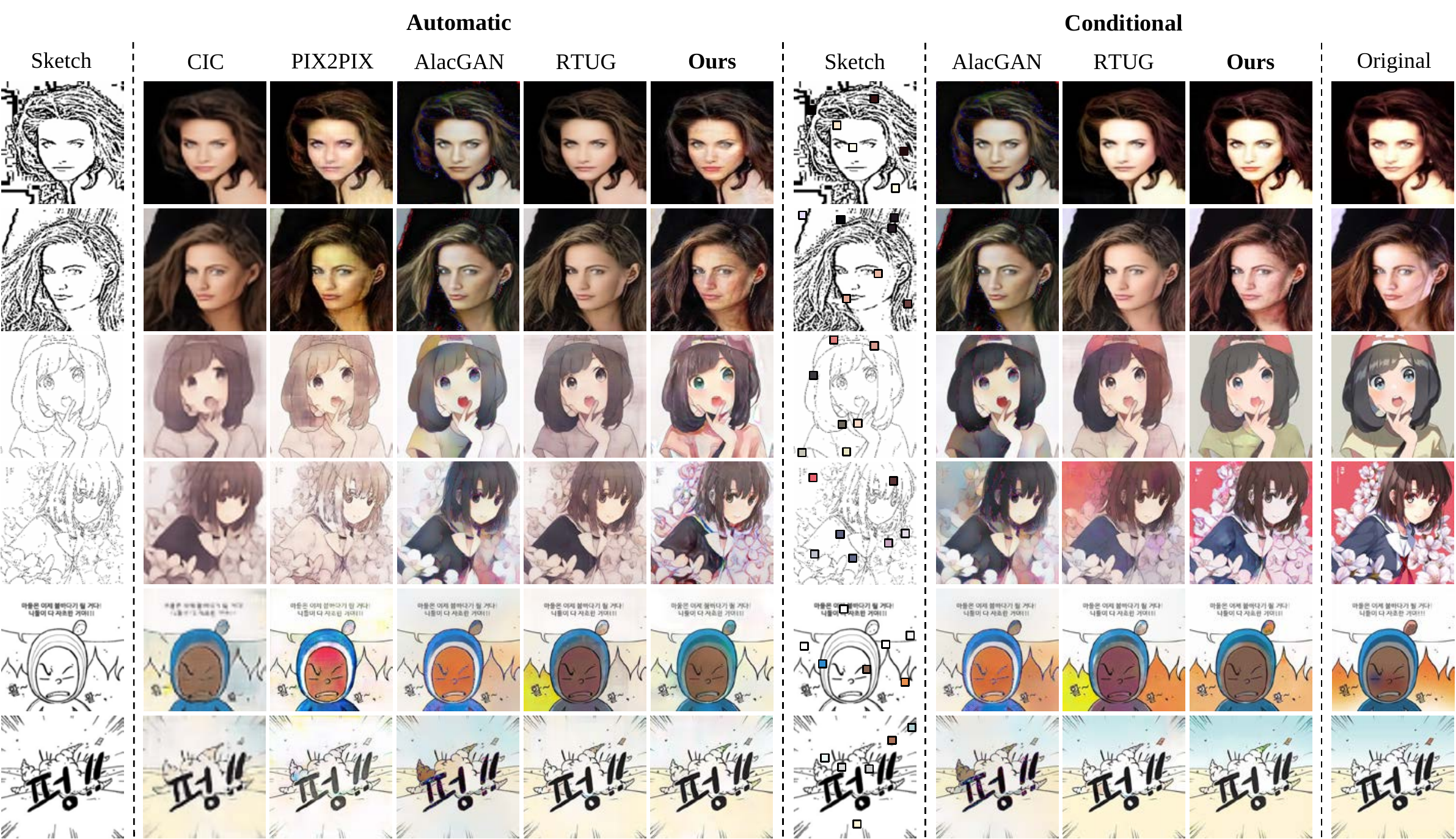}
\caption{\textbf{Comparison to baselines on diverse datasets.} 
We compare our model in two approaches: automatic and conditional colorization ones.
To assess the performance of our model in an automatic setting, we choose CIC, Pix2Pix, AlacGAN, and RTUG as baselines. For the condtional case, we equalize the number of hints given to all baselines and our model. Our model successfully colorizes each segment without color bleeding artifact, e.g., the third and fourth rows, and generates the continuous colors for each segment, e.g., hair in the first two rows, the sky and the ground in the fifth row.
}
\label{fig:qual_comp_full}
\end{figure*}

\begin{figure*}[p]
\begin{center}
\includegraphics[width=\textwidth]{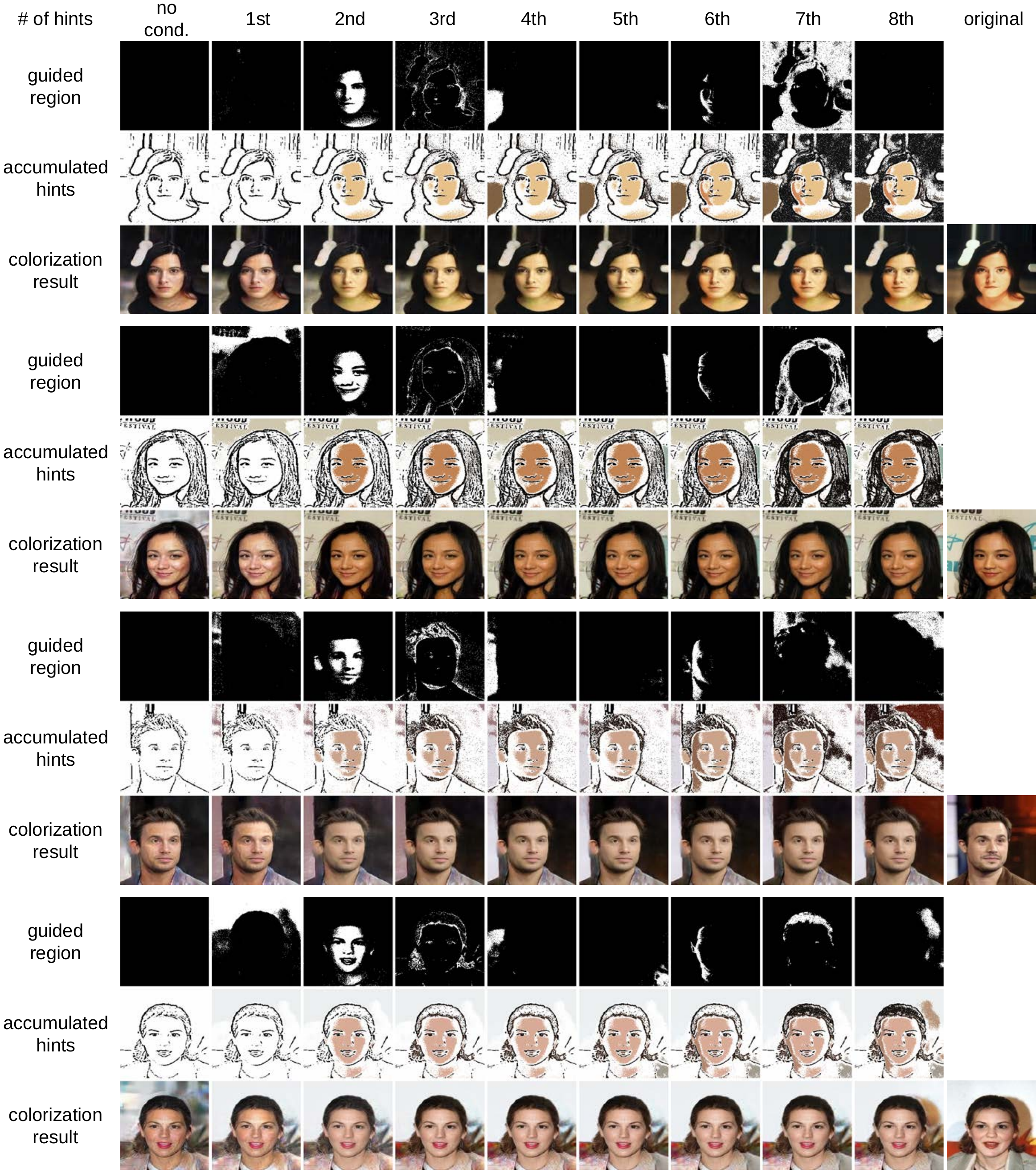}
\end{center}
   \caption{\textbf{Qualitative results on CelebA dataset.} For each interaction, which is noted as number of hints, our model estimates a hint region which the model wants to know first. Then, we select a representative color for the region and the color is spread to the region as visualized in accumulated hints. Finally, a colorization result is generated by our colorization network taking the guided regions accumulated hints and the sketch image. In these examples, we do not remove the noise used in the Gumbel Softmax operation to directly represent the guided regions provided to the colorization network. We summarize the interaction process in three rows for each four image. The results show how the input images change along with the color hints at each interaction step. In particular, the \textit{6-th} column shows that the model captures the shadow of human face and colorizes it appropriate to the image.}
\label{fig:supp_qual1}
\end{figure*}

\begin{figure*}[p]
\begin{center}
\includegraphics[width=\textwidth]{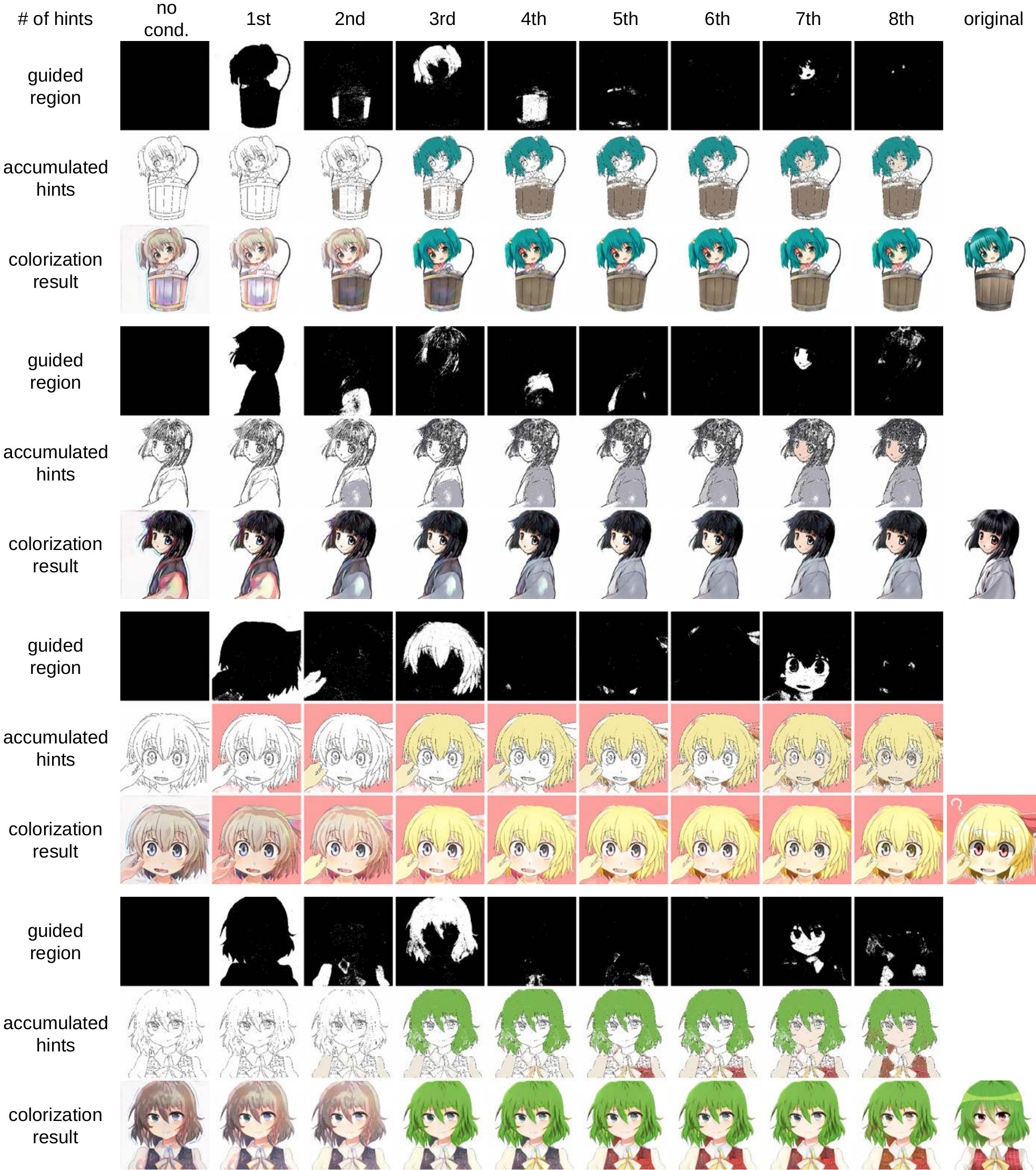}
\end{center}
   \caption{\textbf{Qualitative results on Tag2pix dataset.} Each interaction reveals that our model recognizes semantically related segments for each image, e.g., background, clothes, hair, and face of a character. In the \textit{1-st} iteration, the model concentrates on the background and adapts a color if the background color is inputted. Especially, for the hair segment in the \textit{3-rd} iteration, our model successfully reflects the color changes, not bleeding the color outside of the hair region. }
\label{fig:supp_qual2}
\end{figure*}

\begin{figure*}[p]
\begin{center}
\includegraphics[width=\textwidth]{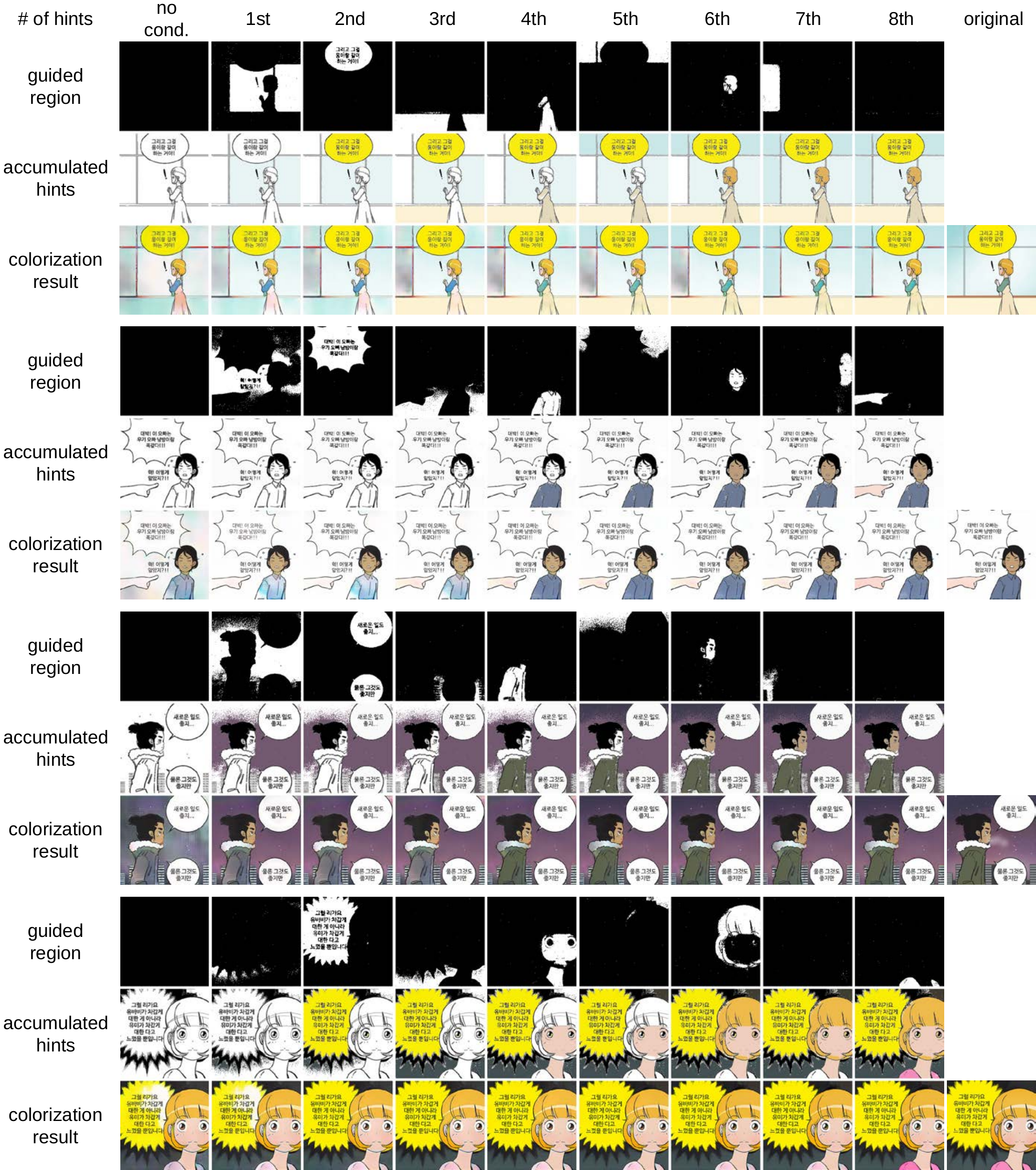}
\end{center}
   \caption{\textbf{Qualitative results on Yumi's Cells dataset.} Each intermediate iteration shows that semantically meaningful segments are recommended and colorized, such as, parts of background, speech balloon, clothes, face, and hair for an image. As shown in the rows of \textit{colorization result}, the automatically colorized images become similar to the groundtruth image by adding each color hint in only eight iteration. This demonstrates that our model not only reflects the color condition in adequate location, but also improve the quality of result images. Especially on the last two images, our model fixes the key color errors, such as the purple night sky, the green clothes, and the yellow speech bubble.}
\label{fig:supp_qual3}
\end{figure*}

\begin{figure*}[p]
\begin{center}
\includegraphics[width=\textwidth]{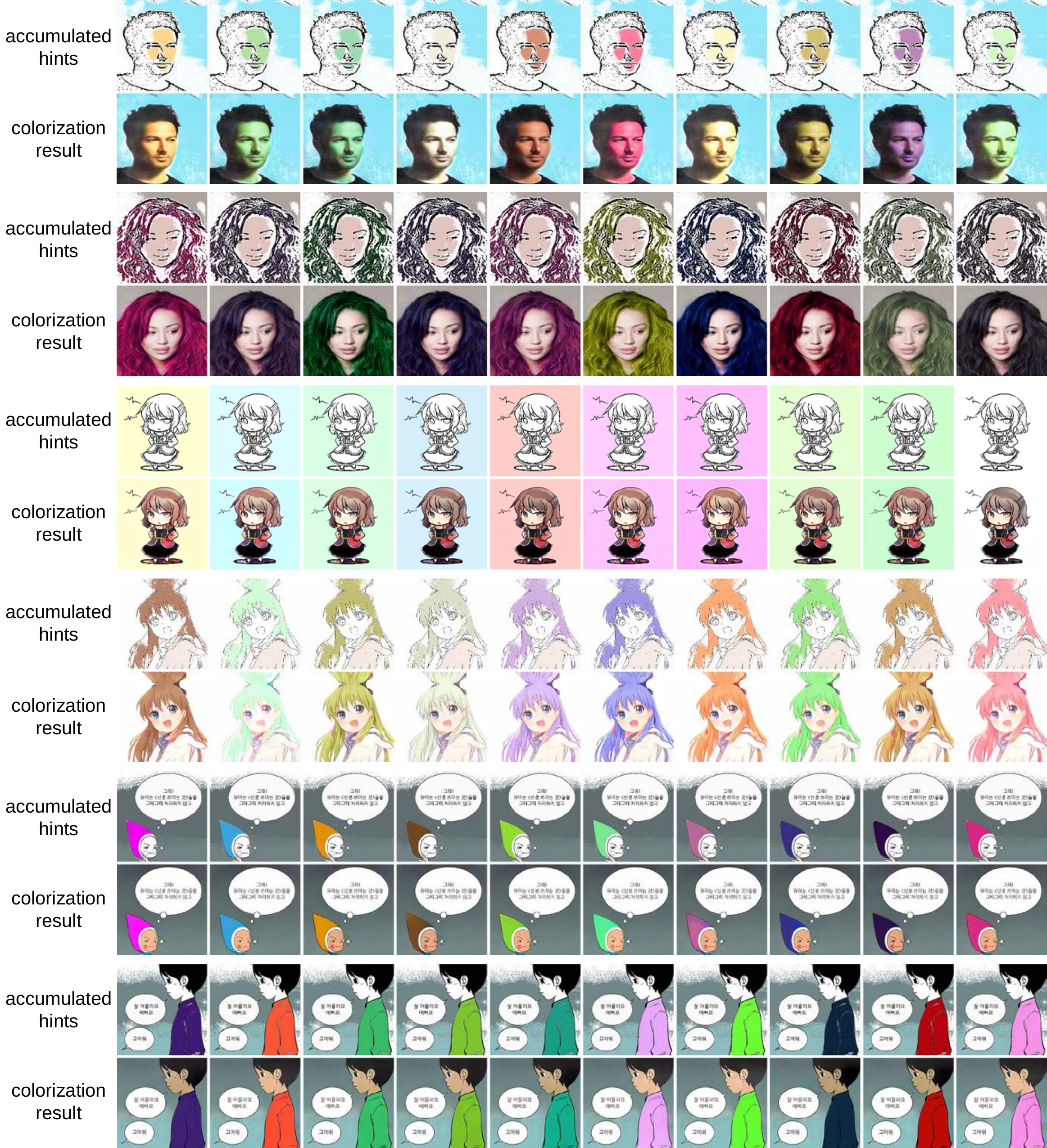}
\end{center}
   \caption{\textbf{Qualitative results of varying color hints on diverse datasets.} We sample two representative items from each three different dataset of CelebA, Tag2pix, and Yumi's cells. To confirm how well our model can reflect user-interaction, we only vary the color of the hint after fixing the input sketch image and its guided regions. The results show our model covers a wide range of color palette, including the green and purple faces.}
\label{fig:supp_qual_per} 
\end{figure*}

This section provides additional qualitative results with the size of $256 \times 256$ over three different datasets.
Fig.~\ref{fig:qual_comp_full} compares qualitative results for both automatic and conditional colorization models.
Fig.~\ref{fig:supp_qual1}, ~\ref{fig:supp_qual2} and ~\ref{fig:supp_qual3} show how the output images approach groundtruth when there are interactions between the model and an user on the CelebA\cite{liu2015celeba}, Tag2pix\cite{Kim2019tag2pix}, and Yumi's Cells\cite{yumicells}, respectively. Fig.~\ref{fig:supp_qual_per} represents diverse output images according to the colors of given hints on each dataset.

\end{document}